\documentclass[11pt,draftcls,onecolumn] {IEEEtran}
\usepackage{graphicx}
\usepackage{epsfig}

\usepackage{amsmath,amssymb,amsfonts}
\usepackage{amsbsy}
\usepackage{graphicx}
\usepackage{graphics}
\usepackage{algorithm}
\usepackage[noend]{algorithmic}
\usepackage{subfigure}
\usepackage{cite}
\usepackage{slashbox}
\usepackage{multirow}
\usepackage{pdfsync}
\usepackage{epstopdf}
\usepackage{color}
\usepackage{amsthm}
\usepackage{boxedminipage}

\newtheorem{proposition}{{Proposition}}

\newcommand{\Rbb}{\mathbb{R}}
\renewcommand{\l}{\ell}

\newcommand{\G}{{\mathcal{G}}}
\newcommand{\E}{{\mathcal{E}}}

\newcommand{\D}{{\mathcal{D}}}
\newcommand{\V}{{\mathcal{V}}}

\renewcommand{\L}{{\mathcal{L}}}
 
 \DeclareMathOperator*{\argmin}{argmin}
\ifCLASSINFOpdf
 
\else
  
\fi

\begin{document}

\title{Learning Parametric Dictionaries for \\Signals on Graphs}
\author{Dorina~Thanou, 
        David~I~Shuman, 
        and Pascal~Frossard       
\thanks{D. Thanou, D. I Shuman and P. Frossard are with Ecole Polytechnique F\'ed\'erale de Lausanne (EPFL), Signal Processing Laboratory-LTS4, CH-1015, Lausanne, Switzerland ({e-mail:\{dorina.thanou, david.shuman, pascal.frossard\}@epfl.ch}).}
\thanks{This work has been partialy funded by the Swiss National Science Foundation under Grant $200021\_135493$.}
\thanks{Part of the work reported here was  presented at the \emph{IEEE Glob. Conf. Signal and Inform. Process., Austin, Texas, Dec. 2013.  }}}
\maketitle

\begin{abstract}
In sparse signal representation, the choice of a dictionary often involves a tradeoff between two desirable properties -- the ability to adapt to specific signal data and a fast implementation of the dictionary. To sparsely represent signals residing on weighted graphs, an additional design challenge is to incorporate the intrinsic geometric structure of the irregular data domain into the atoms of the dictionary. In this work, we propose a parametric dictionary learning algorithm to design data-adapted, structured dictionaries that sparsely represent graph signals. In particular, we model graph signals as combinations of overlapping local patterns. We impose the constraint that each dictionary is a concatenation of subdictionaries, with each subdictionary being a polynomial of the graph Laplacian matrix, representing a single pattern translated to different areas of the graph. The learning algorithm adapts the patterns to a training set of graph signals. Experimental results on both synthetic and real datasets demonstrate that the dictionaries learned by the proposed algorithm are competitive with and often better than unstructured dictionaries learned by state-of-the-art numerical learning algorithms in terms of sparse approximation of graph signals. In contrast to the unstructured dictionaries, however, the dictionaries learned by the proposed algorithm feature localized atoms and can be implemented in a computationally efficient manner in signal processing tasks such as compression, denoising, and classification.

%
%
%In classical  Euclidean settings,  signal decompositions with overcomplete dictionaries offer a way to efficiently approximate or process signals, such that the  important characteristics  are revealed by the sparse signal  representation. In this work, we focus on the sparse representation of graph signals, that are valued-functions living on the vertices of a weighted graph.  In order to identify and exploit structure in the signals, we have clearly to take into account  the intrinsic geometric structure of the underlying weighted graph. We propose a new method for learning overcomplete structured dictionaries that incorporate the graph structure in their components.     We model graph signals as a combination of overlapping    local patterns positioned at different vertices. Then, we impose the constraint that the dictionary is a concatenation of subdictionaries, each describing  a particular signal pattern on the graph. In order to ensure that the atoms are  localized in the graph vertex domain, each subdictionary  is defined by a polynomial of the graph Laplacian. The polynomial kernels permit to include the graph structure in the definition of the atoms, which are also constructed in a computationally efficient way. A novel  algorithm for learning
% the coefficients of the polynomials is then proposed, where constraints related to the well-behaved spectral distribution of the atoms  are included in the optimization problem.  
 \end{abstract}

\IEEEpeerreviewmaketitle

\section{Introduction}
\label{sec:introduction}

\begin{figure*}[t]
\begin{minipage}{2.1in}
\begin{center}
~ \includegraphics[width=1.05\textwidth]{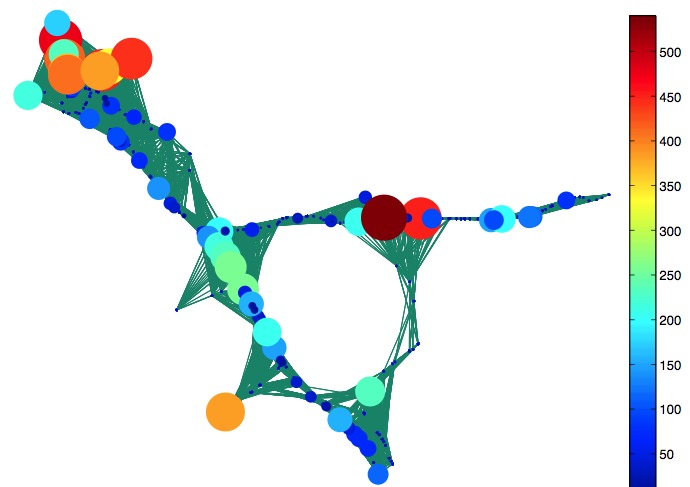}~ \\
~~~(a) Day 1 ~\\
\end{center}
\end{minipage}
\hfill
\begin{minipage}{2.1in}
\begin{center}
~\includegraphics[width=1.05\textwidth]{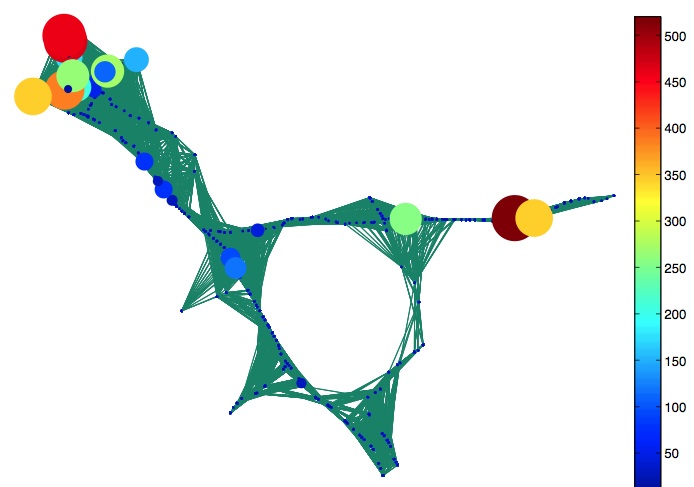}~\\
~~~(b) Day 2~\\
\end{center}
\end{minipage}
\hfill
\begin{minipage}{2.1in}
\begin{center}
~\includegraphics[width=1.05\textwidth]{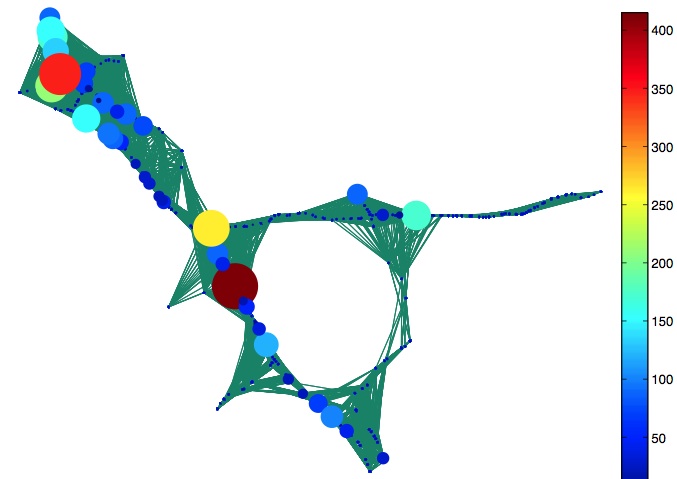}~\\
~~~(c) Day 3 ~\\
\end{center}
\end{minipage}
\caption{Illustrative example:  The three signals on the graph are the minutes of bottlenecks per day  at different detector stations in Alameda County, California, on three different days. The detector stations are the nodes of the graph and the connectivity is defined based on the GPS coordinates of the stations. Note that all signals consist of a set of localized features positioned on different nodes of the graph.}  
\label{fig:illustrative_example}
\vspace{-0.1cm}
\end{figure*}

Graphs are flexible data representation tools, suitable for  modeling the geometric structure of signals that live on topologically complicated domains. Examples of signals residing on such domains can be found in  social, transportation, energy, and sensor networks\cite{Shuman13}. 
In these applications, the vertices of the graph 
represent the discrete data domain,
and the edge weights capture the pairwise relationships between the vertices. A graph signal is then defined as a function that assigns a real value to each vertex. Some simple examples of  graph signals are the current temperature at each location in a sensor network and the traffic level measured at predefined points of the transportation network of a city. An illustrative example is given in Fig. \ref{fig:illustrative_example}.

We are interested in finding meaningful graph signal representations that (i) capture the most important characteristics of the graph signals, and (ii) are sparse. That is, given a weighted graph and a class of signals on that graph, we want to construct an overcomplete dictionary of atoms that can sparsely represent graph signals from the given class as linear combinations of only a few atoms in the dictionary. An additional challenge when designing dictionaries for graph signals is that in order to identify and exploit structure in the data, we need to account for the intrinsic geometric structure of the underlying weighted graph. This is because signal characteristics such as smoothness depend on the topology of the graph on which the signal resides (see, e.g., \cite[Example 1]{Shuman13}).

For signals on Euclidean domains as well as signals on irregular data domains such as graphs, the choice of the dictionary often involves a tradeoff between two desirable properties -- the ability to adapt to specific signal data and a fast implementation of the dictionary \cite{Rubinstein2010overview}. In the \emph{dictionary learning} or \emph{dictionary training} approach to dictionary design, numerical algorithms such as  
K-SVD \cite{Aharon06} 
 and  the Method of Optimal Directions (MOD) \cite{Engan99} 
 (see \cite[Section IV]{Rubinstein2010overview} and references therein) 
 %have been constructed 
 %to 
 learn a dictionary from a set of realizations of the data (training signals). The learned dictionaries are highly adapted to the given class of signals and therefore usually exhibit good representation performance. However, the learned dictionaries are highly non-structured, and therefore costly to apply in various signal processing tasks. On the other hand, analytic dictionaries based on signal transforms such as the Fourier, Gabor, wavelet,  curvelet and shearlet transforms are based on mathematical models of signal classes (see \cite{Mallat2008} and  \cite[Section III]{Rubinstein2010overview}  for a detailed overview of transform-based representations in Euclidean settings). These structured dictionaries often feature fast implementations, but they are not adapted to specific realizations of the data. Therefore, their ability to efficiently represent the data depends on the accuracy of the mathematical model of the data. 
 
The gap between the transform-based representations and  the numerically trained dictionaries  can be bridged by imposing a structure on the dictionary atoms and learning the parameters of this structure. The structure  generally reveals  various desirable properties of the dictionary such as translation invariance \cite{Jost2006}, minimum coherence \cite{Yaghoobi2009} or efficient implementation \cite{Rubinstein2010} (see \cite[Section IV.E]{Rubinstein2010overview} for a complete list of references). Structured dictionaries generally represent a good trade-off between approximation performance and efficiency of the implementation.

In this work, we  build on our previous work \cite{ThanouParamDL} and capitalize on the benefits of both numerical and analytical approaches by learning a dictionary that incorporates the graph structure and can be  implemented efficiently. We model the graph signals as combinations of overlapping local patterns, describing localized events or causes on the graph.  For example, the evolution of  traffic on a highway might be similar to that on a different highway, at a different position in the transportation network. We incorporate the underlying graph structure into the dictionary through the graph Laplacian operator, which encodes the connectivity. In order to ensure the atoms are  localized in the graph vertex domain, we impose the constraint that our dictionary is a concatenation of subdictionaries that are polynomials of the graph Laplacian \cite{Hammond2010}. We then learn the coefficients of the polynomial kernels via numerical optimization. As such, our approach falls into the category of parametric dictionary learning \cite[Section IV.E]{Rubinstein2010overview}. The learned dictionaries are adapted to the training data, efficient to store, and computationally efficient to apply. Experimental results demonstrate the effectiveness of our scheme in the approximation of both synthetic signals and graph signals collected from real world applications.

The structure of the paper is as follows. We first highlight some related work on the representation of graph signals in Section \ref{related_work}.   In Section \ref{preliminary_definitions},  we recall basic definitions related to graphs that are necessary to understand our dictionary learning algorithm.  The polynomial dictionary structure and the dictionary learning algorithms are described in Section \ref{Se:Param_Dict_Learn}. In Section \ref{Se:Exp_results}, we evaluate the performance of our algorithm on the approximation of  both synthetic and real world graph signals. Finally, the benefits of the polynomial structure are discussed in Section \ref{practical_issues}.

\section{Related Work} \label{related_work}
The design of overcomplete dictionaries to sparsely represent signals has been extensively investigated in the past few years. We restrict our focus here to the literature related to the problem of designing dictionaries for graph signals.  Generic numerical approaches such as K-SVD \cite{Aharon06} 
   and MOD \cite{Engan99} can certainly be applied to graph signals, with signals viewed as vectors in $\Rbb^N$. However, the learned dictionaries will neither 
feature a fast implementation, nor  
explicitly incorporate the underlying graph structure. 

Meanwhile, several transform-based dictionaries for graph signals have recently been proposed (see \cite{Shuman13} for an overview and complete list of references). For example, the graph Fourier transform has been shown to sparsely represent smooth graph signals \cite{Zhu12}; wavelet transforms such as diffusion wavelets \cite{Coifman06}, spectral graph wavelets \cite{Hammond2010}, and critically sampled two-channel wavelet filter banks \cite{NarangO12} target piecewise-smooth graph signals; and vertex-frequency frames \cite{ShumanWindGFT,shuman_ACHA_2013,shuman_TSP_2013} can be used to analyze signal content at specific vertex and frequency locations. These dictionaries feature pre-defined structures derived from the graph and some of them can be efficiently implemented; however, they generally are not adapted to the signals at hand. Two exceptions are the diffusion wavelet packets of \cite{bremer_packets} and the wavelets on graphs via deep learning\cite{Rustamov2013}, which feature extra adaptivity.

The recent work in  \cite{Zhang2012} tries to bridge the gap between the graph-based transform methods and the purely numerical dictionary learning algorithms by proposing an algorithm to learn structured graph dictionaries. The learned dictionaries have a structure that is derived from the graph topology, while its parameters are learned from the data. This work is the closest to ours in a sense that both graph dictionaries consist of subdictionaries that are based on the graph Laplacian.  However, it does not necessarily lead to efficient implementations as the obtained dictionary is not necessarily a smooth matrix function (see, e.g., \cite{higham} for more on matrix functions) of the graph Laplacian matrix.

Finally, we remark that the graph structure is taken into consideration in \cite{MiaoZheng2011}, not explicitly into the dictionary but rather in the sparse coding coefficients.  The authors use the graph Laplacian operator as a regularizer in order to impose that the obtained sparse coding coefficients vary smoothly along the geodesics of the manifold that is captured by the graph.
However, the obtained dictionary does not have any particular structure. None of the previous works are able to design dictionaries that provide sparse representations, particularly adapted to a given class of graph signals, and have efficient implementations. This is exactly the objective of our work, where a structured graph signal dictionary is composed of multiple polynomial matrix functions of the graph Laplacian.

%\section{Related Work}
%\subsection{Transform based methods}
%\subsection{Numerical learning techniques}
\section{Preliminaries} \label{preliminary_definitions}
In this section, we briefly overview a few basic definitions for signals on graphs.
A more complete description of the graph signal processing framework can be found in \cite{Shuman13}. 
We consider a weighted and undirected graph $\mathcal{G}=(\V,\E,W)$ where $\V$ and $\E$ represent the vertex and edge sets of the graph, and $W$ represents the matrix of edge weights, with $W_{ij}$ denoting the weight of an edge connecting vertices $i$ and $j$. We assume that the graph is connected.  The  graph Laplacian operator  is defined as $L=D-W$, where $D$ is the diagonal degree matrix  \cite{Chung97}. % \cite{Chung97}.  
The normalized graph Laplacian is defined as $\L=D^{-\frac{1}{2}}LD^{-\frac{1}{2}}$. Throughout the paper, we use the normalized graph Laplacian eigenvectors as the Fourier basis in order to avoid some numerical instabilities that arise when taking large powers of the combinatorial graph Laplacian. Both operators are real symmetric matrices and they have a complete set of orthonormal eigenvectors with corresponding nonnegative eigenvalues. We denote  the eigenvectors  of the normalized graph Laplacian by $\chi=[\chi_1,\chi_2,...,\chi_N]$, and the spectrum of eigenvalues by 
\begin{align*}
\sigma(\L):=\Bigl\{0=\lambda_0<\lambda_1\le\lambda_2\le...\le\lambda_{N-1}\leq 2\Bigr\}.
\end{align*} %The $j$ hop neighborhood $\mathcal{N}_{j,n}=\{v\in V: d(v,n)\le j\}$ of node $n$ is the set of all nodes which  are at most $j$- hop distance away from node $n$.
%\footnote{%The framework would be similar in the case of the unnormalized Laplacian. 
%In general, we prefer to work with the normalized Laplacian in order to avoid dealing with computational  issues when we assume large powers of the polynomial.}

A graph signal $y$ in the vertex domain is a real-valued %scalar 
function %$y\in \Rbb^N$ 
defined on the vertices of the graph $\G$, such that $y(v)$ is the %sample 
value of the function at vertex $v \in \V$. The spectral domain representation can also provide significant information about the characteristics of %the graph 
graph signals. In particular, the eigenvectors of the Laplacian operators can be used to perform %a 
 harmonic analysis of signals that live on the graph, and  the corresponding eigenvalues %represent the graph frequencies
 carry a notion of frequency \cite{Shuman13}. The normalized Laplacian eigenvectors are a  Fourier basis, so that for any   function $y$ defined on the vertices of the graph, the graph Fourier transform $\hat{y}$ at frequency $\lambda_{\l}$ is defined as 
%a function of eigenvectors with
\begin{equation}
\small{
\nonumber \hat{y}\left(\lambda_{\l}\right)=\langle y,\chi_{\ell} \rangle=\sum_{n=1}^N y(n)\chi^*_{\ell}(n), }
\end{equation}
while the inverse graph Fourier transform is 
\begin{equation}
\small{
\nonumber y(n)=\sum_{{\ell}=0}^{N-1}\hat{y}\left(\lambda_{\ell}\right) \chi_{\ell}(n), \quad \forall n\in \V.
}
\end{equation}

%\subsection{Translation and Localization on the Graph}
Besides its use in harmonic analysis, the  graph Fourier transform is also useful in defining the translation of a signal on the graph. 
The generalized translation operator can be defined as a generalized convolution with a Kronecker $\delta$ function centered at vertex $n$ \cite{ShumanWindGFT,shuman_ACHA_2013}:
%Given a kernel $g$ defined in the vertex domain, the translation operator $T_n: \mathbb{R}^N\to\mathbb{R}^N$ at a vertex $n$ is defined as the generalized convolution of the kernel $g$ with a unit impulse function ($\delta$ function)  centered at vertex $n$ \cite{Shuman13}:%\cite{ShumanWindGFT}:
\begin{equation}
\small{
T_ng=\sqrt{N}(g*\delta_n)=\sqrt{N}\sum_{{\ell}=0}^{N-1}\hat{g}(\lambda_{\ell})\chi_{\ell}^{*}(n)\chi_{\ell}.
\label{translation}
}
\end{equation}
The right-hand side of \eqref{translation} allows us to interpret the generalized translation as an 
 %The translation on graphs can be interpreted as 
 %an 
 operator acting on the kernel $\hat{g}(\cdot)$, which is defined directly in the graph spectral domain, and the
%\subsection{Localization in the vertex domain}
 %The 
 localization of $T_n g$ around the center vertex $n$ is controlled by the smoothness of the kernel $\hat{g}(\cdot)$ \cite{Hammond2010,shuman_ACHA_2013}.  %More precisely, for a smooth kernel $g$, the magnitude of $T_ng(m)$ decays as the distance between the nodes $n$ and $m$ increases. 
 One can thus design atoms $T_n g$ that are localized around $n$ in the vertex domain by taking the kernel $\hat{g}(\cdot)$ in \eqref{translation} to be a  smooth polynomial function of degree $K$:  %integer powers of the Laplacian can be used\cite{Hammond2010}. 
\begin{equation}
\small{
\hat{g}(\lambda_{\ell})=\sum_{k=0}^K\alpha_{k}\lambda_{\ell}^{k}, \quad \ell=0,...,N-1.
}
\label{kernels}
\end{equation}
%By c
Combining (\ref{translation}) and (\ref{kernels}), we can translate a polynomial kernel to each vertex in the graph and generate a set of $N$ localized atoms, which are the columns of %the corresponding translated kernels in all the vertices of the graph are given by 
\begin{equation}
Tg=\sqrt{N}\hat{g}(\L)=\sqrt{N}\chi \hat{g}(\Lambda)\chi^T=\sqrt{N}\sum_{k=0}^K\alpha_{k}\L^{k},
\end{equation}
where $\Lambda$ is the diagonal matrix of the eigenvalues. Note that the $k^{th}$ power of the Laplacian $\L$ is exactly $k$-hop localized on the graph topology \cite{Hammond2010}; i.e.,  for the atom centered on node $n$, if there is no $K$-hop path connecting nodes $n$ and $i$, then $(T_n g)(i)=0$ . This definition of localization is based on the shortest path distance on the graph and ignores the weights of the edges.

\section{Parametric dictionary learning on graphs}\label{Se:Param_Dict_Learn}
Given a set of training signals on a weighted graph, our objective is to learn a structured dictionary that sparsely represents classes of graph signals. We consider a general class of graph signals that are linear combinations of (overlapping) graph patterns positioned at different vertices on the graph.  We aim to learn a dictionary that is capable of capturing all possible translations of a set of patterns. We use the definition \eqref{translation} of generalized translation, and we learn a set of polynomial generating kernels (i.e., patterns) of the form \eqref{kernels} that capture the main characteristics of the signals in the spectral domain. Learning directly in the spectral domain enables us to detect spectral components that exist in our training signals, such as atoms that are supported on selected frequency components.   In this section,  we describe in detail the structure of our dictionary  and the learning algorithm.

\subsection{Dictionary Structure} \label{Se:structure}
We design a structured graph dictionary $\D=[\D_1,\D_2,...,\D_S]$ that is a concatenation of a set of $S$ subdictionaries of the form
\begin{equation}
\small{
\D_s=\widehat{g_s}(\L)=\chi\left(\sum_{k=0}^K\alpha_{sk}\Lambda^{k}\right)\chi^{T}=\sum_{k=0}^K\alpha_{sk}{\L}^{k},
\label{subdictionary}
}
\end{equation}
where $\widehat{g_s}(\cdot)$ is the generating kernel or pattern of the subdictionary $\D_s$. 
%The block structure permits us to represent each generating kernel and its corresponding atoms separately.  In particular,  the structure per block  leads to an easy interpretation of the atoms in term of their position on the graph.  
Note that the atom given by column $n$ of subdictionary $\D_s$ is equal to $\frac{1}{\sqrt{N}}T_n g_s$; i.e., the  polynomial $\widehat{g_s}(\cdot)$ of order $K$ translated to the vertex $n$. The polynomial structure of the kernel $\widehat{g_s}(\cdot)$ ensures that the resulting atom given by column $n$ of subdictionary $\D_s$ has its support contained in a $K$-hop neighborhood of vertex $n$ \cite[Lemma 5.2]{Hammond2010}.  

%Since each atom is itself a  signal on the graph, the constraint on the support of the atom is consistent with  the assumption that the value of a signal at each vertex is usually affected by the value of a signal in a small neighborhood of the graph.  

The polynomial constraint guarantees the localization of the atoms in the vertex domain, but it does not provide any information about the spectral representation of the atoms. In order to control their frequency behavior,  we  impose two constraints on the spectral representation of the kernels $\left\{\widehat{g_s}(\cdot)\right\}_{s=1,2,\ldots,S}$. First, we require that the kernels are nonnegative and uniformly bounded by a given constant $c$. In other words,  we impose that  $0 \leq \widehat{g_s}(\lambda)\leq c$ for all $\lambda \in [0,\lambda_{\max}]$, or, equivalently,
 \begin{equation} \label{Eq:constraint1}
 \small{
0 \preceq \D_s	\preceq cI,~~\forall s\in\{1,2,...,S\},
}
\end{equation}
where $I$ is the $N \times N$ identity matrix. 
Each subdictionary $\D_s$ has to be a positive semi-definite matrix whose maximum eigenvalue is upper bounded by $c$. 

Second, since the classes of signals under consideration usually contain frequency components that are spread across the entire spectrum, %  In order to cover all the spectrum through the learned kernels,
the learned kernels $\left\{\widehat{g_s}(\cdot)\right\}_{s=1,2,\ldots,S}$ should also cover the full spectrum.  %One way to ensure this would be to impose a constraint that $\sum_{s=1}^S |\widehat{g_s}(\lambda)|^2$ is constant for all $\lambda \in [0,\lambda_{\max}]$, which, following a slight generalization of \cite[Theorem 5.6]{Hammond2010}, also guarantees that the resulting dictionary $\D$ is a tight frame.
%However, such a constraint would lead to a significantly more difficult optimization problem for the learning phase discussed in the next subsection.
We thus 
impose the constraint 
 \begin{equation}
 \small{
c-\epsilon_1\le \sum_{s=1}^S\widehat{g_s}(\lambda)	\le c+\epsilon_2, \quad \forall \lambda\in[0, \lambda_{\max}],
}
\end{equation} 
 or equivalently   
 \begin{equation} \label{Eq:constraint2}
\small{
 (c-\epsilon_1)I \preceq \sum_{s=1}^S \D_s	\preceq (c+\epsilon_2)I,}
\end{equation} 
  where  $\epsilon_1, \epsilon_2$ are  small positive constants. Note that both (\ref{Eq:constraint1}) and (\ref{Eq:constraint2}) are quite generic and do not assume any particular prior on the spectral behavior of the atoms. If we have additional  prior information, we can incorporate that prior into our optimization problem by modifying these constraints.    For example, if we know that our signals' frequency content is restricted to certain parts of the spectrum, by choosing $\epsilon_1$ close to $c$,  we allow the flexibility for our learning algorithm to learn filters covering only these parts and not  the entire spectrum. %we can guide the algorithm to learn kernels that are concentrated only on a particular band and do not cover the whole spectrum.  
  
  Finally,  the spectral constraints  increase the stability of the dictionary. From the constants $c$, $\epsilon_1$ and $ \epsilon_2$, we can derive frame bounds for $\D$, as shown in the following proposition.  
  %The above relaxation, however, leads to a dictionary that is a frame and the frame bounds are computed based on the following proposition. 
  \begin{proposition} 
  Consider a dictionary  $\D=[\D_1,\D_2,...,\D_S]$, where each $\D_s$ is of the form of $\D_s=\sum_{k=0}^K\alpha_{sk}{\L}^{k}$. If the kernels $\left\{\widehat{g_s}(\cdot)\right\}_{s=1,2,\ldots,S}$ satisfy the constraints 
  $0 \leq \widehat{g_s}(\lambda)\leq c$  and $ c-\epsilon_1\le \sum_{s=1}^S\widehat{g_s}(\lambda)	\le c+\epsilon_2,$ for all $\lambda \in [0,\lambda_{\max}]$
then the set of atoms $\left\{{d_{s,n}}\right\}_{s=1,2,\ldots,S, n=1,2,...,N}$ of $\D$ form a frame. For every signal $y\in \Rbb^N$,
  \begin{equation}\label{frame}
  \nonumber \frac{(c-\epsilon_1)^2}{S}\|y\|^2_2\le\sum_{n=1}^N\sum_{s=1}^{S}|\langle y,d_{s,n}\rangle |^2\le (c+\epsilon_2)^2\|y\|^2_2.
  \end{equation}
%  \begin{equation}\label{frame}
%  \nonumber A\|y\|^2_2\le\sum_{n=1}^N\sum_{s=1}^{S}|\langle y,d_{s,n}\rangle |^2\le B\|y\|^2_2,
%  \end{equation}
%   where  $\frac{(c-\epsilon_1)^2}{S}\le A:=\min\{\sum_{s=1}^{S}| \widehat{g_s}(\lambda_{\ell}) |^2\}$ and 
%   $ B:=\max\{\sum_{s=1}^{S}| \widehat{g_s}(\lambda_{\ell}) |^2\}\le (c+\epsilon_2)^2.$ 
   \begin{proof}
  % The proof follows from a  generalization of  \cite[Theorem 5.6]{Hammond2010}.
From \cite[Lemma 1]{shuman_TSP_2013}, which is a slight generalization of \cite[Theorem 5.6]{Hammond2010}, we have 
   \begin{align}
 \label{Parseval}  
 \sum_{n=1}^N\sum_{s=1}^{S}|\langle y,d_{s,n}\rangle |^2=
 \sum_{\ell=0}^{N-1}| \hat{y}(\lambda_{\ell})|^2\sum_{s=1}^{S}| \widehat{g_s}(\lambda_{\ell}) |^2,~~ \forall \lambda \in \sigma(\L). 
% &=\sum_{n=1}^N\sum_{s=1}^{S}|\langle \hat{y},\widehat{d_{s,n}}\rangle |^2\\ 
%%\nonumber &=\sum_{n=1}^N\sum_{s=1}^{S}|\langle \hat{y},\widehat{g_s(\L)(:,n)}\rangle |^2\\
%\nonumber&=\sum_{n=1}^N\sum_{s=1}^{S}\sum_{\ell=0}^{N-1}| \hat{y}(\lambda_{\ell})|^2|\widehat{d_{s,n}}(\lambda_{\ell}) |^2\\
%\label{def_atom}&=\sum_{\ell=0}^{N-1}\sum_{n=1}^N\sum_{s=1}^{S}| \hat{y}(\lambda_{\ell})|^2|\chi_{\ell}(n)|^2| \widehat{g_s}(\lambda_{\ell}) |^2\\
%\nonumber&=\sum_{\ell=0}^{N-1}| \hat{y}(\lambda_{\ell})|^2\sum_{n=1}^N|\chi_{\ell}(n)|^2\sum_{s=1}^{S}| \widehat{g_s}(\lambda_{\ell}) |^2\\
%&\label{square_eq}=\sum_{\ell=0}^{N-1}| \hat{y}(\lambda_{\ell})|^2\sum_{s=1}^{S}| \widehat{g_s}(\lambda_{\ell}) |^2
   \end{align}
%   where (\ref{Parseval}) is due to Parseval's theorem, and (\ref{def_atom}) follows from the structure of the dictionary. 
 From the constraints  on the spectrum of kernels  $\left\{\widehat{g_s}(\cdot)\right\}_{s=1,2,\ldots,S}$ we have 
   \begin{equation}
 \label{Beta}  \sum_{s=1}^{S}| \widehat{g_s}(\lambda_{\ell}) |^2\le \left(\sum_{s=1}^{S} \widehat{g_s}(\lambda_{\ell}) \right)^2\le(c+\epsilon_2)^2,~~\forall \lambda \in \sigma(\L) 
   \end{equation}
   Moreover, from the left side of (\ref{Eq:constraint2}) and the Cauchy-Schwarz inequality, we have
   \begin{equation}
\label{Alpha}\frac{(c-\epsilon_1)^2}{S}\le \frac{ \left(\sum_{s=1}^{S} \widehat{g_s}(\lambda_{\ell}) \right)^2}{S}  \le \sum_{s=1}^{S}| \widehat{g_s}(\lambda_{\ell}) |^2,~~\forall \lambda \in \sigma(\L).
   \end{equation}
   Combining (\ref{Parseval}), (\ref{Beta}) and  (\ref{Alpha}) yields the desired result.
   %, for $y\ne 0$ we obtain
   %\begin{equation}\label{frame}
  %\nonumber A\|y\|^2_2\le\sum_{n=1}^N\sum_{s=1}^{S}|\langle y,d_{s,n}\rangle |^2\le B\|y\|^2_2.
  %\end{equation}
   \end{proof}
  
  \end{proposition}
 %One way to ensure this would be to impose a constraint that $\sum_{s=1}^S |\widehat{g_s}(\lambda)|^2$ is constant for all $\lambda \in [0,\lambda_{\max}]$, which, following a slight generalization of \cite[Theorem 5.6]{Hammond2010}, also guarantees that the resulting dictionary $\D$ is a tight frame.
We remark that if we alternatively impose that  $\sum_{s=1}^S |\widehat{g_s}(\lambda)|^2$ is constant for all $\lambda \in [0,\lambda_{\max}]$, 
the resulting dictionary  $\D$ would be  a tight frame. However, such a constraint leads to a significantly more difficult optimization problem to learn the dictionary. The constraints \eqref{Eq:constraint1} and \eqref{Eq:constraint2} %on    the parameters $\left\{\alpha_{sk}\right\}_{s=1,2,\ldots,S;~k=1,2,\ldots,K}$,% that are  constrained by \eqref{Eq:constraint1} and \eqref{Eq:constraint2} 
%for stability  issues and
lead to an easier dictionary learning optimization problem, while still 
 providing control on the spectral representation of the atoms and the stability of signal reconstruction with the learned dictionary. 
 
 To summarize, the dictionary $\D$ is a parametric dictionary that depends on the parameters 
 \begin{align*}
 \left\{\alpha_{sk}\right\}_{s=1,2,\ldots,S;~k=1,2,\ldots,K},
 \end{align*}
 and the constraints \eqref{Eq:constraint1} and \eqref{Eq:constraint2} can be viewed as constraints on these parameters.

\subsection{Dictionary Learning Algorithm}\label{Se:dictionary_learning_alg}
Given a set of training signals $Y=[y_1,y_2,...,y_M]\in \Rbb^{N\times M}$, all living on the weighted graph $\mathcal{G}$, our objective is to learn a graph dictionary $\D\in \Rbb^{N\times NS}$ with the structure described in Section \ref{Se:structure} that can efficiently represent all of the signals in $Y$ as linear combinations of only a few of its atoms. Since $\D$ has the form \eqref{subdictionary}, this is equivalent to learning the parameters $\left\{\alpha_{sk}\right\}_{s=1,2,\ldots,S;~k=1,2,\ldots,K}$ that characterize the set of generating kernels, $\left\{\widehat{g_s}(\cdot)\right\}_{s=1,2,\ldots,S}$. We denote these parameters in vector form as $\alpha=[\alpha_1;...;\alpha_S]$, where $\alpha_s$ is a column vector with $(K+1)$ entries.

Therefore, the dictionary learning problem can be cast as the following optimization problem:
%{\small
\begin{align}\label{eq:opt_prob}
\argmin_{\alpha \in \Rbb^{(K+1)S} ,~X \in \Rbb^{SN \times M}} & \left\{ || Y - \D X ||^{2}_{F}+\mu \|\alpha\|_2^2 \right\}  \\ 
 \mbox{subject to~~~~~~}  & \|x_m\|_0 \le T_0, \quad \forall m\in\{1,...,M\}, \nonumber \\ 
  &   \D_s=\sum_{k=0}^K \alpha_{sk}\L^k, \quad \forall s\in\{1,2,...,S\} \nonumber \\
 &   0\preceq \D_s	\preceq c, \quad \forall s\in\{1,2,...,S\} \nonumber \\
 &  (c-\epsilon_1)I\preceq \sum_{s=1}^S\D_s	\preceq (c+\epsilon_2)I, \nonumber
\end{align} %}
%
%\begin{equation}
%{\small
%\begin{split}
% &\underset{\alpha \in \Rbb^{(K+1)S} ,~X \in \Rbb^{SN \times M}}{\operatorname{argmin}} \left\{ || Y - D X ||^{2}_{F}+\mu \|\alpha\|^2 \right\}
% \\ &   \mbox{   subject to }   \|x_m\|_0\le T_0 \quad \forall m\in\{1,...,M\},\\ &   \quad  \quad  \quad  \quad \quad0\preceq D_i	\preceq c, \forall i\in\{1,2,...,S\}\\&
%\quad  \quad  \quad  \quad \quad c-\epsilon\preceq \sum_{i=1}^SD_i	\preceq c+\epsilon,\\
%\end{split}
%\label{eq:opt_prob}
%}
%\end{equation}
%\hspace{-.05in}
where $\D=[\D_1,\D_2,\ldots,\D_S]$, $x_m$ corresponds to column $m$ of the %atom 
coefficient matrix $X$, and  $T_0$ is the sparsity level of the coefficients of each signal. 
%The solution $\alpha^{*}$ represents the best set of parameters in terms of sparse approximation of the training signals $Y$. 
Note that in the objective of the optimization problem \eqref{eq:opt_prob}, 
%optimization problem 
we penalize the norm of the polynomial coefficients $\alpha$ in order to (i) promote smoothness in the learned polynomial kernels, and (ii) improve the numerical stability of the learning algorithm.

The optimization problem (\ref{eq:opt_prob}) is not %jointly 
convex, but it can be approximately solved  in a computationally efficient manner 
by  alternating between the sparse coding and %the 
dictionary update steps. In the first step, we fix the parameters $\alpha$ (and accordingly fix the dictionary $\D$ via the structure \eqref{subdictionary}) and solve 
\begin{equation}
\small{
\nonumber\underset{ X}{\operatorname{argmin}}|| Y - \D X ||^{2}_{F} \quad  \mbox{ subject to } \|x_m\|_0\le T_0 \quad \forall m\in\{1,...,M\},
}
\end{equation}
using orthogonal matching pursuit (OMP) \cite{Tropp04}, \cite{Bruckstein2009}, which has been shown to perform well in the dictionary learning literature. Before applying OMP, we normalize the atoms of the dictionary so that they all have a unit norm.  This step  is essential for the OMP algorithm in order to  treat all of the atoms equally. After computing the coefficients $X$, we renormalize the atoms of our dictionary  to recover our initial polynomial structure \cite[Chapter 3.1.4]{Elad2010} and  the sparse coding coefficients in such a way that the product $\D X$ remains constant.   Note that other methods for solving the sparse coding step such as  matching pursuit (MP) or iterative soft thresholding could be used as well. %However, in the case of iterative soft thresholding, since the atoms of the dictionary do not have a unit norm,  a careful tuning of the step size is needed. 

In the second step, we fix the coefficients $X$ and update the dictionary %$\D$ 
by finding the vector of %coefficients
parameters, $\alpha$, that solves %minimizes the following:
%{\small
\begin{align}\label{eq:opt_prob_fix_X}
 \argmin_{\alpha \in \Rbb^{(K+1)S}} & \left\{ || Y - \D X ||^{2}_{F}+\mu \|\alpha\|_2^2 \right\}  \\ 
 \mbox{subject to~~}  
&  \D_s=\sum_{k=0}^K \alpha_{sk}\L^k, \quad \forall s\in\{1,2,...,S\} \nonumber \\
 & 0\preceq \D_s	\preceq cI, \quad \forall s\in\{1,2,...,S\} \nonumber \\ 
 &  (c-\epsilon_1)I\preceq \sum_{s=1}^S \D_s	\preceq (c+\epsilon_2)I. \nonumber
\end{align}
%}
%\hspace{-.05in}Note that $\alpha$ determines $\D_s$ and $\D$ according to \eqref{subdictionary}. 
Problem \eqref{eq:opt_prob_fix_X} is convex and can be written as a quadratic program. %\cite{BoydConvex}. 
The details are given in the Appendix \ref{QP}. Algorithm \ref{Alg_DL} contains a summary of the basic steps of our dictionary learning algorithm.

Since the optimization problem (\ref{eq:opt_prob}) is solved by alternating between the two steps, the polynomial dictionary learning algorithm is not guaranteed to converge to the optimal solution; however, the algorithm converged in all of experiments to a local optimum.   
%At each iteration, we first update the  sparse coding coefficients for a fixed sparsity level and then update the dictionary parameters by keeping the sparse coding coefficients fixed.  Converge to a local minimum is then guaranteed under the constraint that the sparse coding step is solved successfully. In our experiments, we have observed that this is always the case.  
Finally, the complexity of our algorithm is %mainly 
dominated by the pre-computation of the eigenvalues of the Laplacian matrix, which we use to enforce %needed for 
the spectral constraints on the kernels. In the dictionary update step, the quadratic program (line 10 of Algorithm \ref{Alg_DL}) can be efficiently solved in polynomial time using %classical 
optimization techniques such as %conventional 
interior point methods  \cite{BoydConvex}  or %the more recently introduced 
operator splitting methods  (e.g., Alternating Direction Method of Multipliers \cite{Boyd_admm}).  The former methods lead to more accurate solutions, while the latter 
are better suited to solve large scale problems. For the numerical examples in this paper, we use interior point methods to solve the quadratic optimization problem.
%are characterized by a higher accuracy in the solution of the optimization problem while the latter are well-suited to solve large scale problems at the cost of  lower precision.     In the rest of this work, we use interior point methods for solving the quadratic optimization problem. However, we have observed in practice that both methods lead to similar results.      %On the other hand, the polynomial structure can reduce the complexity of the sparse coding step, as discussed in Section \ref{Computational_issues}. 
 \begin{algorithm}[htb]
\caption{Parametric  Dictionary Learning on Graphs}
\begin{algorithmic}
\item [ 1:] {\bf Input:} Signal set $Y$, initial dictionary $\D^{(0)}$, target signal sparsity $T_0$, polynomial degree $K$, number of subdictionaries $S$, number of iterations $iter$
\item [ 2:] {\bf Output:}  Sparse signal representations $X$, polynomial coefficients $\alpha$
\item [ 3:] {\bf Initialization:} $\D=\D^{(0)}$
\item [ 4:] {{\bfseries for} $i=1,2,...,iter$ \bfseries{do}:}
\item [ 5:]  \quad {\bfseries Sparse Approximation Step:} 
\item [ 6:]  \quad \quad (a) Scale  each atom in $\D$ to a unit norm
\item [ 7:]  \quad \quad (b) Solve $\nonumber\underset{ X}{\operatorname{min}}|| Y - \D X ||^{2}_{F} \quad  \mbox{ subject to } \|x_m\|_0\le T_0 \quad \forall m\in\{1,...,M\}$
\item [ 8:]  \quad \quad (c) Rescale $X$ and $\D$ to recover the polynomial structure
\item [ 9:]  \quad {\bfseries Dictionary Update Step:} 
\item [10:]  \quad \quad Compute the polynomial coefficients $\alpha$ by solving the optimization problem (\ref{eq:opt_prob_fix_X}), and update the dictionary according to \eqref{subdictionary}
%\item [ 5:] Compute and store $c^{1}_n=(\mathcal{L}^Ty)_n$.
%\item [ 6:]{{\bfseries for} $l=2,...,k$ \bfseries{do}:}
%\item [ 7:]  \quad Transmit $c^{l-1}_n=(\mathcal{L}^Tc^{l-2})_n$ to all the neighbors 
%\item [ 8:]  \quad Receive $c^{l-1}_m$ from all the neighbors $m\in \mathcal{N}_n$.
%\item [ 9:] {{\bfseries end for}} 
%\item [10:] {\bfseries for} ${i=1,..,S}$ {\bfseries do} 
%\item [11:] \quad Compute $(\D^Ty)_{(i-1)N+n}=\alpha_{0i}y(n)+\sum_{l=1}^k \alpha_{li}c^l_n  $
\item [11:] {\bf end for}  
\end{algorithmic}
\label{Alg_DL}
\end{algorithm}

\section{Experimental results} \label{Se:Exp_results}
In the following experiments, we quantify the performance of the proposed dictionary learning method in the approximation of  both synthetic and real data. First, we study the behavior of our algorithm in the synthetic scenario where the signals are linear combinations of  a few localized atoms that are placed on different vertices of the graph. Then, we study the performance of our algorithm in the approximation of graph signals collected from real world applications. In all  experiments,  we compare the performance  of our algorithm  to the performance of (i) graph-based transform methods such as the spectral graph wavelet transform (SGWT)\cite{Hammond2010}, (ii) purely numerical dictionary learning methods such as K-SVD \cite{Aharon06} that treat the graph signals as vectors in $\Rbb^N$ and ignore %completely 
the graph structure, and (iii)  the graph-based dictionary learning algorithm presented in \cite{Zhang2012}. The kernel bounds in (\ref{eq:opt_prob}), if not otherwise specified,  are chosen as $c=1$ and $\epsilon_1=\epsilon_2=0.01$, and the number of iterations in the learning algorithm is fixed to 25.  We use the \emph{sdpt3} solver  \cite{SDPT3}  in the  \emph{yalmip} optimization toolbox \cite{YALMIP} %in order 
to solve the quadratic problem  \eqref{eq:opt_prob_fix_X} in the learning algorithm. %\footnote{YALMIP is  publicly available at http://users.isy.liu.se/johanl/yalmip/} 
In order to directly compare the methods mentioned above, we always use orthogonal matching pursuit (OMP) for  the sparse coding step in the testing phase, where we first normalize the dictionary atoms to a unit norm. We could alternatively apply %an 
iterative soft thresholding %for computing 
to compute the sparse coding coefficients; however, that would require a careful tuning of the stepsize for each of the algorithms. Finally, the average approximation error is set to $\|Y_{test}-\D X_{test}\|_F^2/|Y_{test}|$, where $|Y_{test}|$ is the size of the testing set and $X_{test}$ are the sparse coding coefficients. % for each of the tested dictionaries.

\subsection{Synthetic Signals}
We first study the performance of our algorithm for the approximation of synthetic signals. We generate a graph by randomly placing  $N=100$ vertices  in the unit square.   We set the edge weights   based on a thresholded Gaussian kernel function so that $ W(i,j)=e^{-\frac{[dist(i,j)]^2}{2\theta^2}}$ if the physical distance between vertices $i$ and $j$ is less than or equal to $\kappa$, and zero otherwise.
%, \mbox { if }  dist(i,j) \le \kappa$, and 0 otherwise. 
We fix $\theta=0.9$ and $\kappa=0.5$ in our experiments, and ensure that the graph is connected. 

\subsubsection{Polynomial Generating Dictionary}
\begin{figure}[h]
      \centering
            \subfigure[Kernels of the generating dictionary]{ \includegraphics[width=7cm]{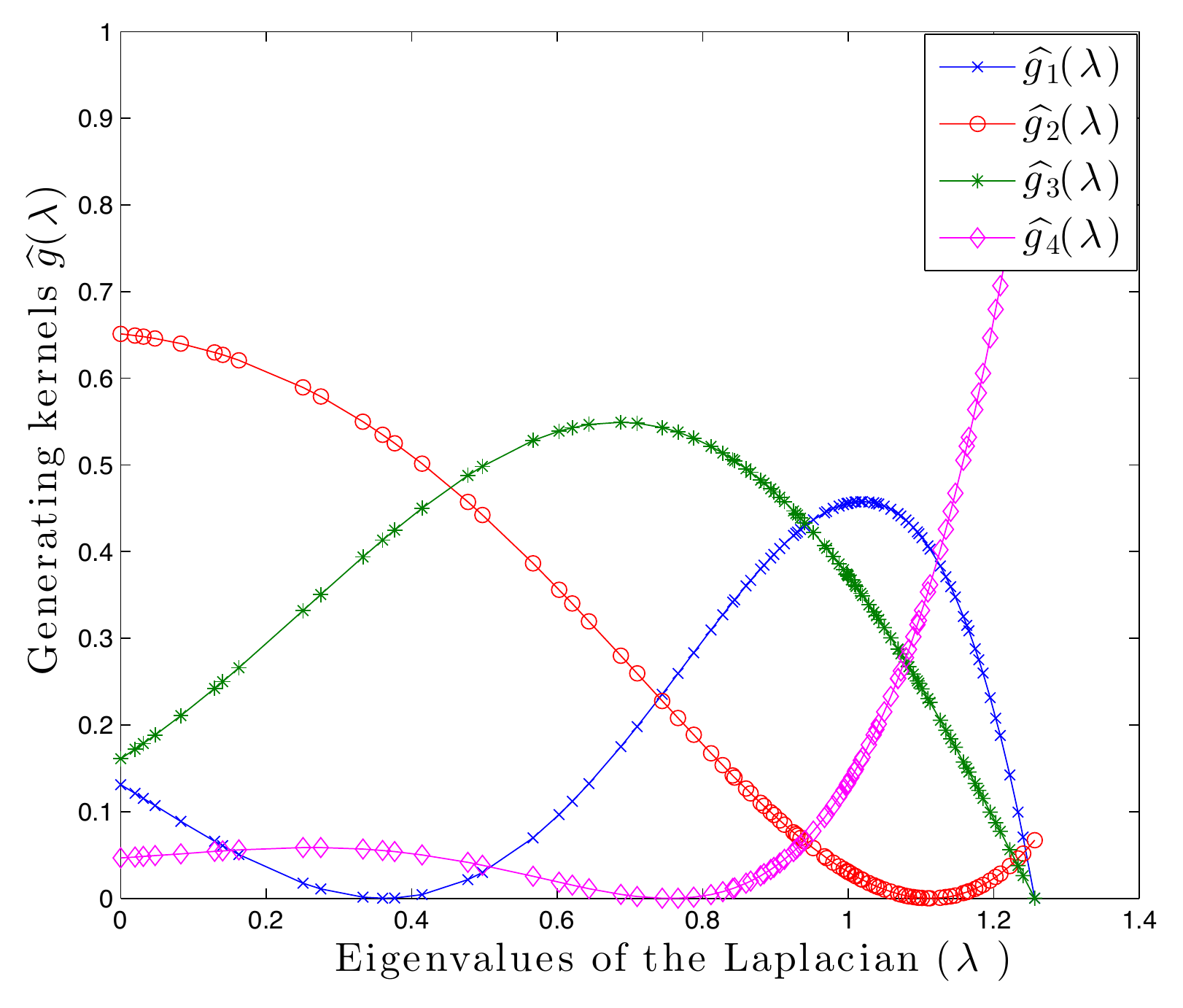}   \label{original_kernels}}
            \subfigure[Learned kernels with $M=400$]{ \includegraphics[width=7cm]{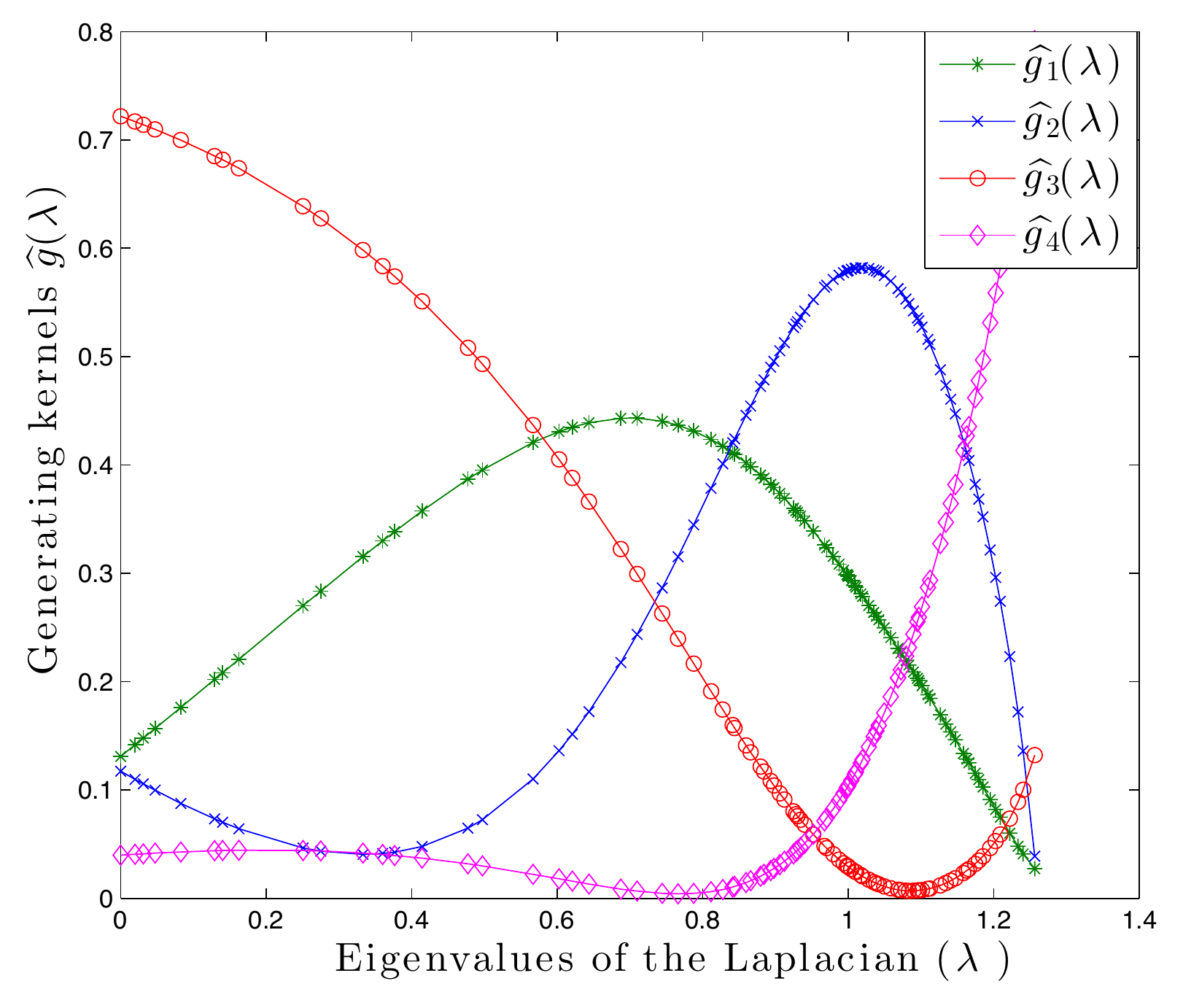} \label{kernels400}}
               \subfigure[Learned kernels with $M=600$]{ \includegraphics[width=7cm]{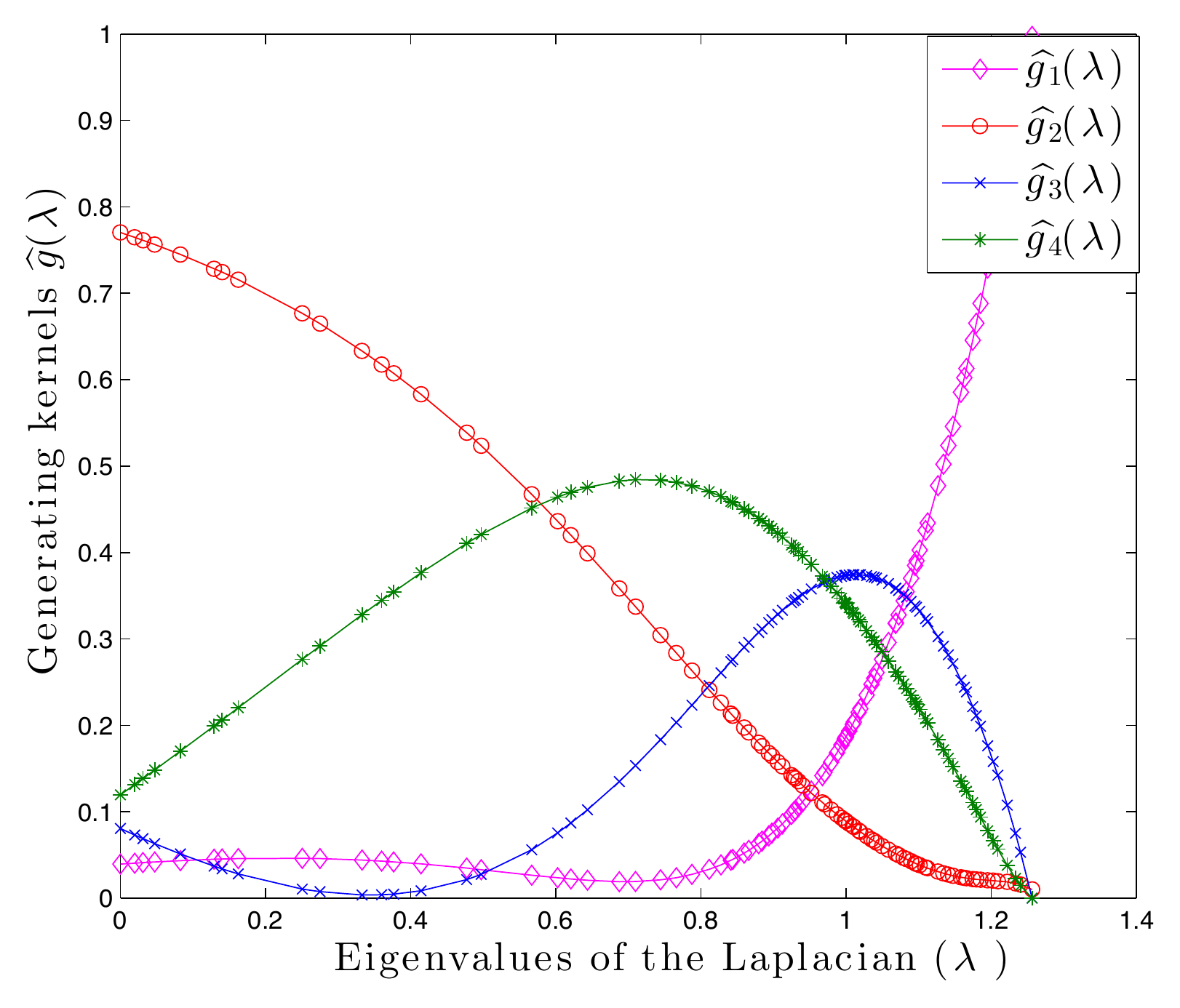}\label{kernels600}}
                  \subfigure[Learned kernels with $M=2000$]{ \includegraphics[width=7cm]{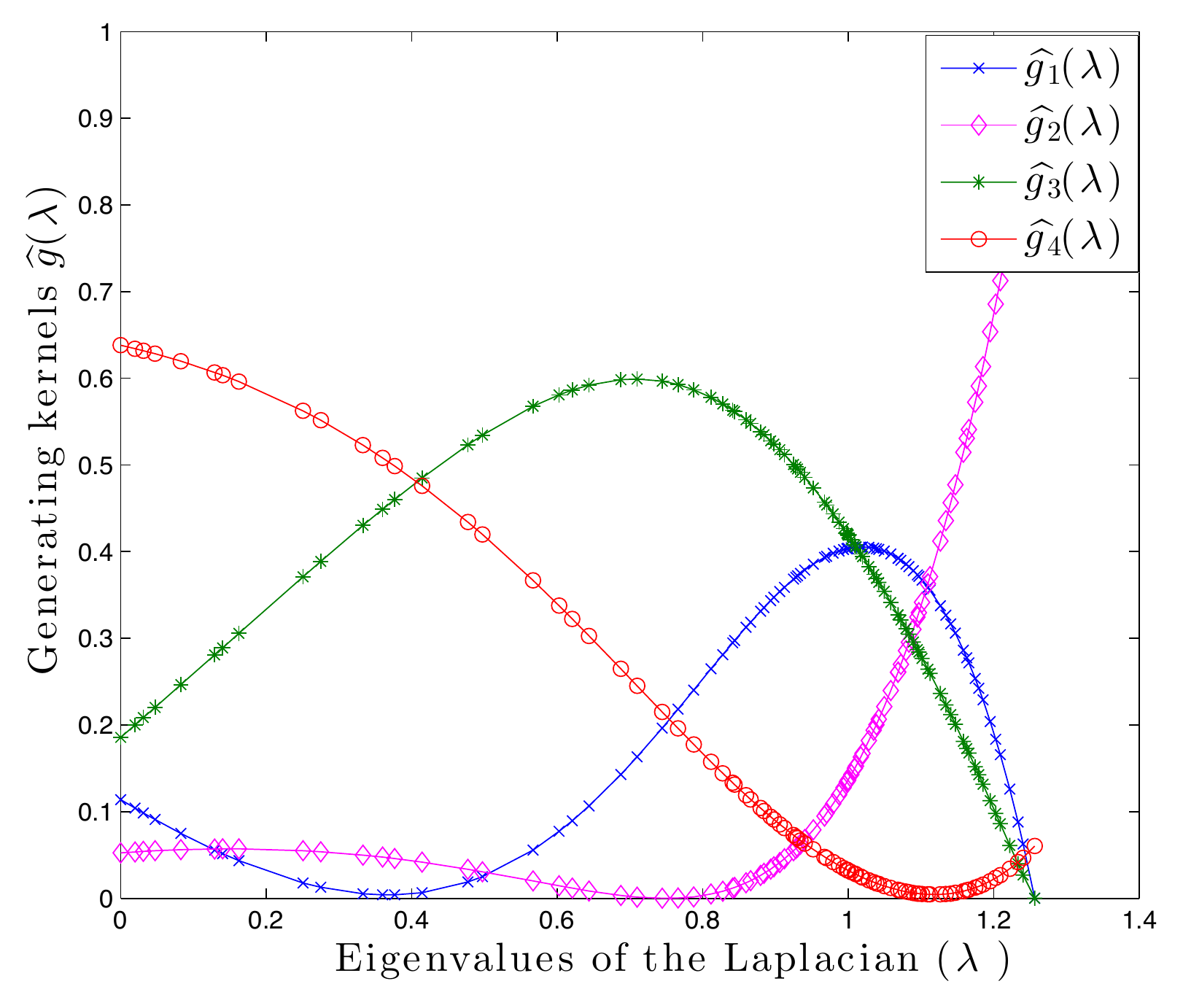}\label{kernels2000}}
        \caption{Comparison of the kernels learned by the polynomial dictionary learning algorithm to the generating kernels $\left\{\widehat{g_s}(\cdot)\right\}_{s=1,2,\ldots,S}$ (shown in (a)) for $M=400$,  $M=600$ and  $M=2000$ training signals.} 
        \label{fig:difMkernels}
%\vspace{-0.5cm}
\end{figure}

In our first set of experiments, to construct a set of synthetic training signals consisting of localized patterns on the graph, we use a generating dictionary that is a concatenation of $S=4$ subdictionaries.  Each subdictionary is a fifth order ($K=5$) polynomial of the graph Laplacian according to (\ref{subdictionary}) and  captures one of the four constitutive components of our signal class. The generating kernels  $\left\{\widehat{g_s}(\cdot)\right\}_{s=1,2,\ldots,S}$ of the dictionary are shown in Fig. \ref{original_kernels}. 
We generate the graph signals by linearly combining  $T_0\le4$ random atoms from the dictionary with random coefficients.  We then learn a dictionary from the training signals, and we expect this learned dictionary to be close to the known generating dictionary. % that is used for creating the synthetic signals.   % In the following experiments, we study  how different parameters affect the performance of the polynomial graph dictionary.  
%\subsubsection{Size of the training set} \label{TrainingSetExp1} 

We first study the influence of the size of the training set on the dictionary learning outcome. %This is an important parameter in learning algorithms as training a big dictionary usually requires a large number of training signals in the training phase.  However, 
Collecting a large  number of training signals can be infeasible in many applications. Moreover, training a dictionary with a large training set significantly increases the complexity of the learning phase, leading to impractical optimization problems.  %Hence, we  examine now the stability of the learned  dictionary with respect to the size of the training signals. 
Using our polynomial dictionary learning algorithm with training sets of $M=\{400, 600, 2000\}$ signals, we learn a dictionary of $S=4$ subdictionaries. To allow some flexibility into our learning algorithm, we fix the degree of the learned polynomials to $K=20$. 
%We first study the recovery performance of our  algorithm with respect to the number of training signals. In Fig. \ref{fig:difMkernels} we illustrate the obtained kernels. 
Comparing Fig. \ref{original_kernels} to Figs. \ref{kernels400}, \ref{kernels600}, and \ref{kernels2000}, we observe that our algorithm is able to recover the shape of the kernels used for the generating dictionary, even with a very small number of training signals.  However, the accuracy of the recovery improves as we increase the size of the training set.  To quantify the improvement, we define the mean SNR of the learned kernels as $\frac{1}{S}\sum_{s=1}^S-20\log(\|\widehat{g_s}(\Lambda)-\widehat{g_s}^{'}(\Lambda)\|_2)$, where $\widehat{g_s}(\Lambda)$ is the true pattern of Fig.  \ref{original_kernels} for the subdictionary $\D_s$ and    $\widehat{g_s}^{'}(\Lambda)$ is the corresponding  pattern learned with our polynomial dictionary algorithm. The SNR values that we obtain are $\{4.9, 5.3, 14.9\}$ for  $M=\{400, 600, 2000\}$, respectively.
%, which indicate that, as we increase the size of the training set the recovery performance of the algorithm improves. In all the cases, the learned atoms are pretty close to the groundtruth atoms, which confirms the efficiency of the learning algorithm in this first idealistic scenario.   %verify the dependence of our dictionary learning ability of our dictionary learning algorithm to recover the spectral band  imposed in the synthetic dictionary and how it is related to the maximum degree $K$ of the polynomials.

%\begin{figure*}[h]
%\begin{minipage}{2.1in}
%\begin{center}
%~ \includegraphics[width=1.1\textwidth]{Figures/kernelsM400K20}~ \\
%~~~(a) M=400 ~\\
%\end{center}
%\end{minipage}
%\hfill
%\begin{minipage}{2.1in}
%\begin{center}
%~\includegraphics[width=1.1\textwidth]{Figures/kernelsM600K20}~\\
%~~~(b) M=600 ~\\
%\end{center}
%\end{minipage}
%\hfill
%\begin{minipage}{2.1in}
%\begin{center}
%~\includegraphics[width=1.1\textwidth]{Figures/kernelsM2000K20}~\\
%~~~(c) M=2000 ~\\
%\end{center}
%\end{minipage}
%\caption{Learned kernels with the polynomial dictionary algorithm  with $K=20$ for a training set of size (a) $M=400$, (b) $M=600$ and (c) $M=2000$.  }  
%\label{fig:difMkernels}
%\vspace{-0.1cm}
%\end{figure*}

\begin{figure*}[t]
\begin{minipage}{2.1in}
\begin{center}
~ \includegraphics[width=1.05\textwidth]{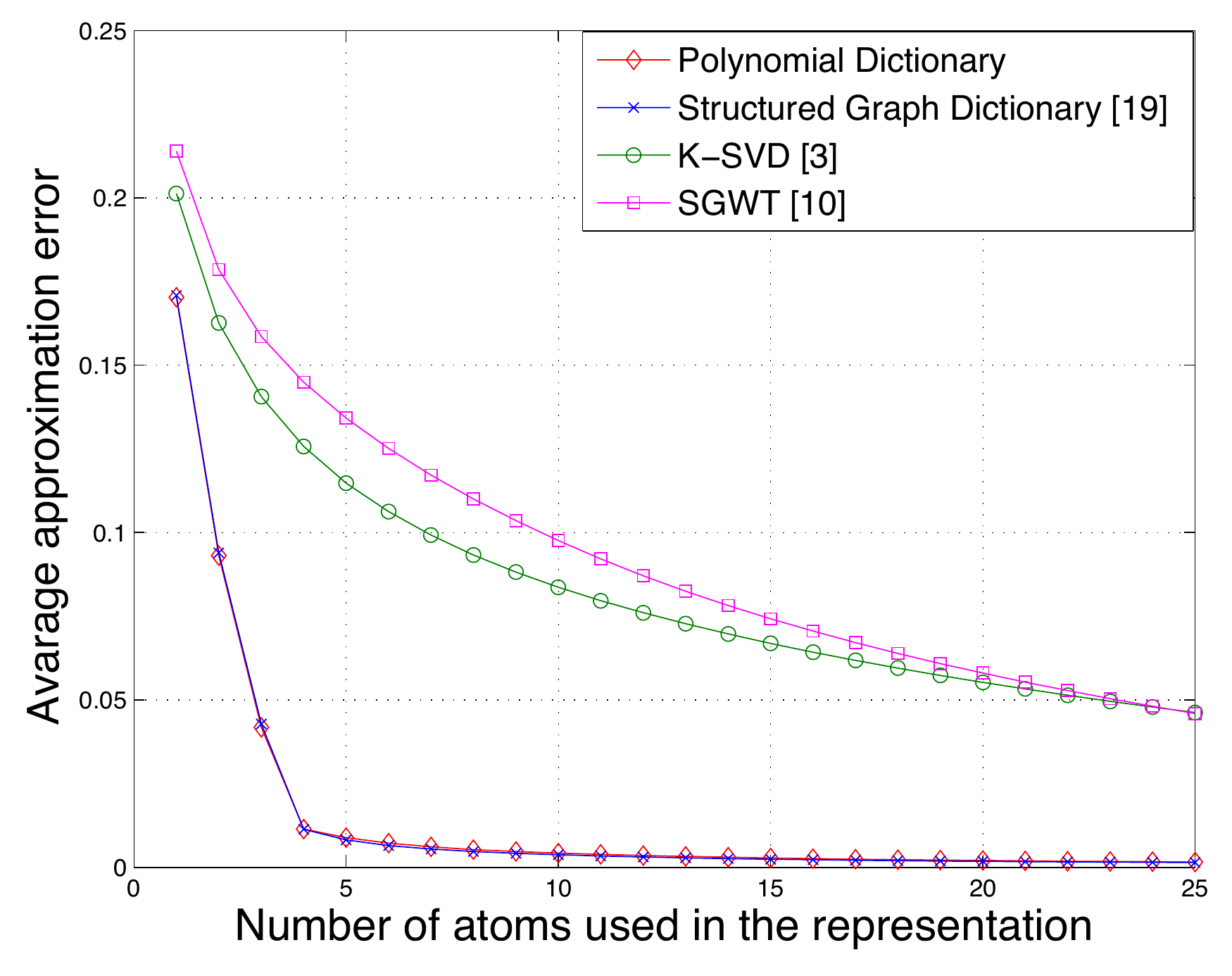}~ \\
~~~(a) M=400 ~\\
\end{center}
\end{minipage}
\hfill
\begin{minipage}{2.1in}
\begin{center}
~\includegraphics[width=1.05\textwidth]{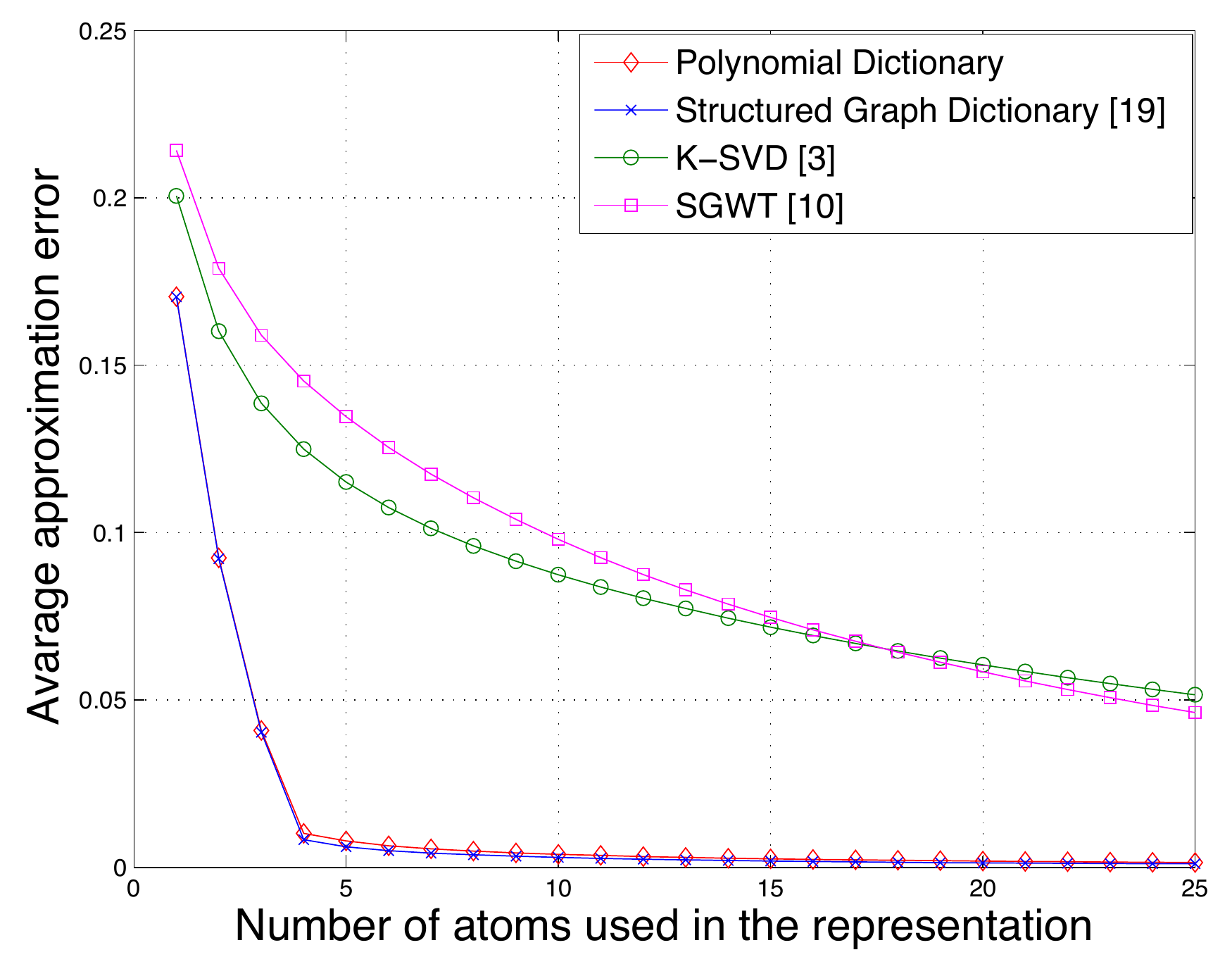}~\\
~~~(b) M=600 ~\\
\end{center}
\end{minipage}
\hfill
\begin{minipage}{2.1in}
\begin{center}
~\includegraphics[width=1.05\textwidth]{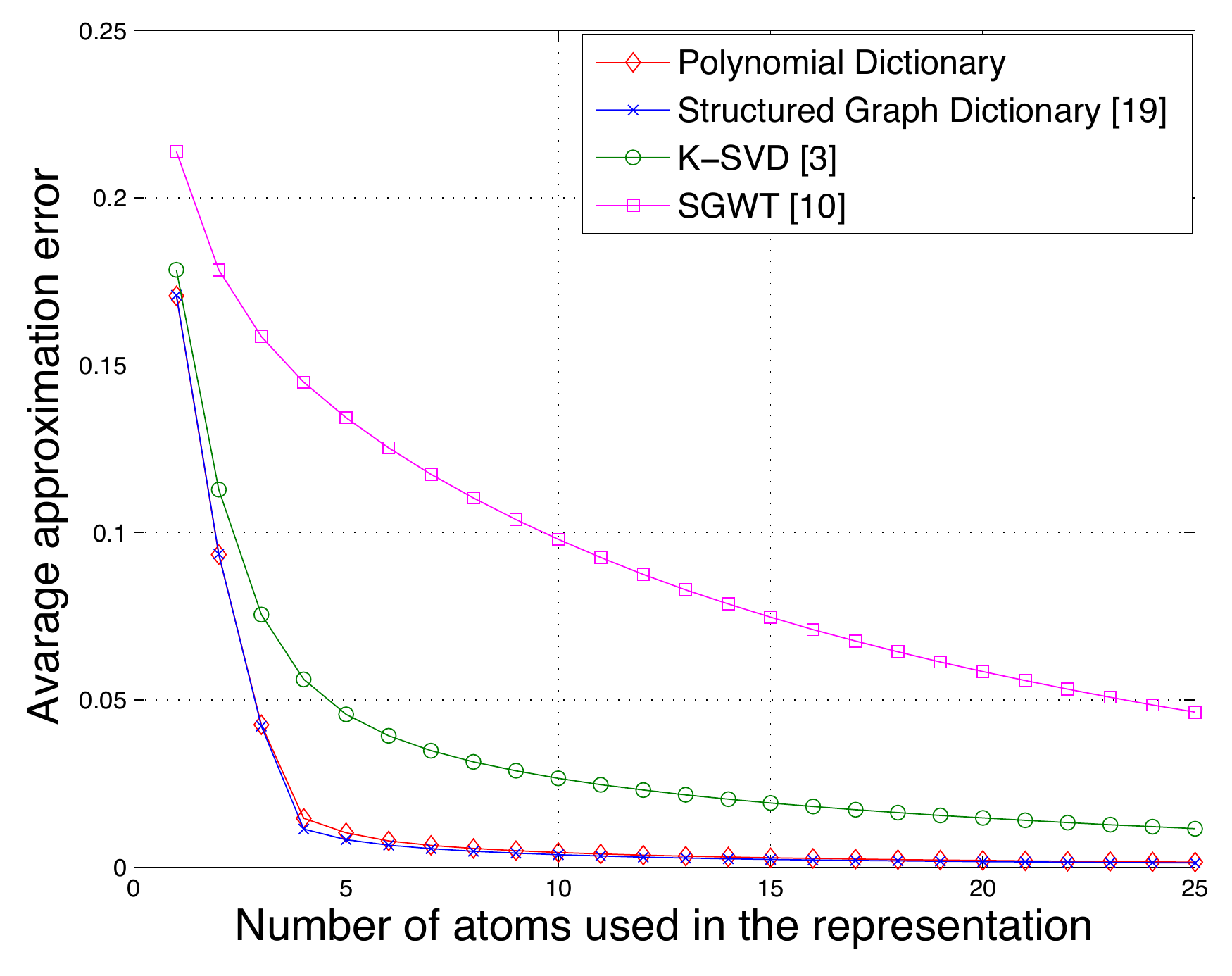}~\\
~~~(c) M=2000 ~\\
\end{center}
\end{minipage}
\caption{Comparison of the learned polynomial dictionary to the SGWT\cite{Hammond2010}, K-SVD \cite{Aharon06} and the graph structured dictionary \cite{Zhang2012} in terms of approximation performance on test data generated from a polynomial generating dictionary, for different sizes of the training set.}  
\label{fig:difM1}
\vspace{-0.1cm}
\end{figure*}

Next, we generate 2000 testing signals using the same method as for the the construction of the training signals. We then study the effect of the size of the training set on the approximation of the testing signals with atoms from our learned dictionary.     %Moreover, we compare the performance of our scheme with the one obtained with the SGWT\cite{Hammond2010}, K-SVD \cite{Aharon06} and the graph structured dictionary \cite{Zhang2012}. 
Fig. \ref{fig:difM1} illustrates the results for three different sizes of the training set and compares the approximation performance to that of other learning algorithms. Each point in the figure is the average of 20 random runs with different realizations of the training and testing sets. We first observe that the approximation performance of the polynomial dictionary is always better than that of SGWT, which demonstrates the benefits of the learning process. The improvement is attributed to the fact that the SGWT kernels are designed \emph{a priori}, while our algorithm learns the shape of the kernels from the data.  

\begin{figure*}[t]
\begin{minipage}{2.1in}
\begin{center}
~ \includegraphics[width=1.2\textwidth]{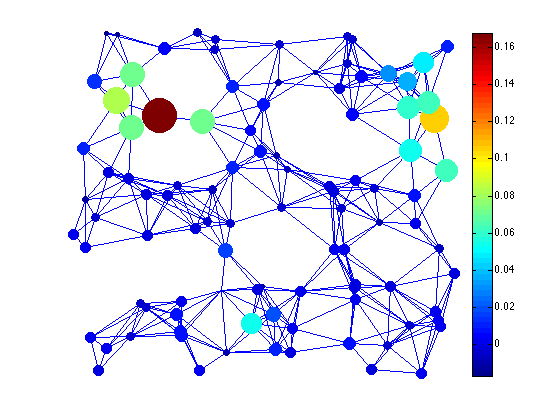}~ \\
~~~(a) Graph signal~\\
\end{center}
\end{minipage}
\hfill
\begin{minipage}{2.1in}
\begin{center}
~\includegraphics[width=1.2\textwidth]{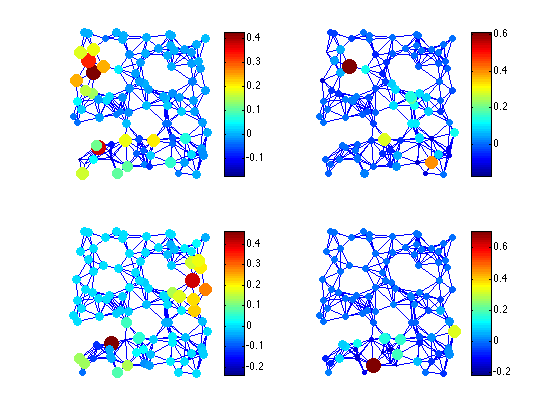}~\\
~~~(b)  K-SVD ~\\
\end{center}
\end{minipage}
\hfill
\begin{minipage}{2.1in}
\begin{center}
~\includegraphics[width=1.2\textwidth]{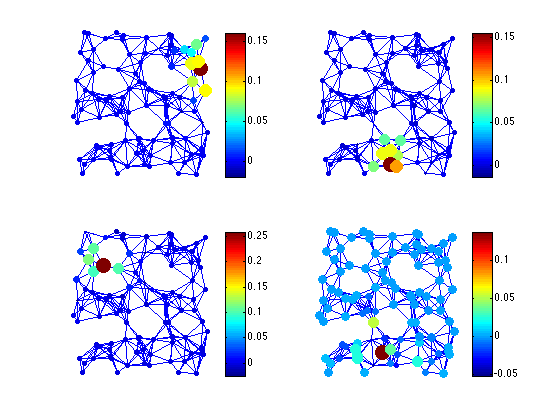}~\\
~~~(c) Polynomial dictionary  ~\\
\end{center}
\end{minipage}
\caption{(a)   An example of a graph signal from the testing set and its  atomic decomposition with respect to (b) the dictionary learned by K-SVD and (c) the learned polynomial graph dictionary.   Note that the K-SVD atoms have a more global support while the polynomial dictionary atoms are localized in specific neighborhoods of the graph. }  
\label{fig:signal_decomposition_example}
\vspace{-0.1cm}
\end{figure*}

We also see that the performance of K-SVD  depends on the size of the training set. Recall that K-SVD is blind to the graph structure, and is therefore unable to capture translations of similar patterns. In particular, we observe that when the size of the training set is relatively small, as in the case of $M=\{500, 600\}$, the approximation performance of K-SVD significantly deteriorates. It  slightly improves when the number of training signals increases (i.e., $M=2000$). Our polynomial dictionary however shows much more stable performance with respect to the size of the training set.  We note two reasons that may underly the better performance of our algorithm, as compared to K-SVD. First, K-SVD tends to learn atoms that sparsely approximate the signal on the whole graph, rather than to extract common features that appear in different neighborhoods. As a result, the atoms learned by K-SVD tend to have a global support on the graph, and K-SVD shows poor performance in the datasets  containing many localized signals. An example of the atomic decomposition of a graph signal from the testing set with respect to the K-SVD and the polynomial graph dictionary is illustrated in Fig. \ref{fig:signal_decomposition_example}. %We observe that indeed the selected atoms with OMP from the K-SVD dictionary have a more global support in the graph, while the decomposition with respect to our dictionary consists of localized features.  
Second, even when K-SVD does learn a localized pattern appearing in the training data, it only learns that pattern in that specific area of the graph, and does not assume that similar patterns may appear at other areas of the graph. Of course, as we increase the number of training signals, translated instances of the pattern are more to likely appear in other areas of the graph in the training data, and K-SVD is then more likely to learn atoms containing such patterns in different areas of the graph. On the other hand, our polynomial dictionary learning algorithm learns the patterns in the graph spectral domain, and then includes translated versions of the patterns to all locations in the graph in the learned dictionary, even if some specific instances of the translated patterns do not appear in the training data.

%Since it ignores the graph  structure, the K-SVD dictionary is not translation invariant  thus it is quite sensitive to the actual signals  in the training phase. Therefore,   a testing signal that contains a translated pattern  that is not included in any of the training signals, can not be well approximated with the K-SVD dictionary. This  often happens when the  size of the training signal set  is small, where the performance of K-SVD is extremely poor. However, as we increase significantly the number of training signals, the translated instances of the patterns  over the graph will most probably appear in the training phase and thus the K-SVD dictionary will exhibit a good performance on the validation set. On the other hand, with our polynomial dictionary learning algorithm, we learn directly the patterns in the spectral domain and we are able to obtain translated versions of the pattern over the graph, even if some specific instances of the translated patterns are missing during the training phase.  
    %From the practical point of view, the K-SVD dictionary is highly non-structured and quite complex to apply. %These results illustrate the tradeoff  between the adaptability and the complexity of a dictionary. 
    
 The algorithm proposed in \cite{Zhang2012} represents some sort of intermediate solution between K-SVD and our algorithm. It learns a dictionary that  consists of subdictionaries of the form $\chi \widehat{g_s}(\Lambda)\chi^T$, where the specific values $\widehat{g_s}(\lambda_0), \widehat{g_s}(\lambda_1), \ldots,\widehat{g_s}(\lambda_{N-1})$ are learned, rather than learning a continuous kernel $\widehat{g_s}(\cdot)$ and evaluating it at the $N$ discrete eigenvalues as we do. %$\widehat{g_s}$ does not follow any particular function model, but rather consists of some discrete values.  
As a result, the obtained dictionary is adapted to the graph structure and it contains atoms that are translated versions of the same pattern on the graph. 
%Due to the graph structure, the performance of the algorithm is stable with respect to the size of the training set as well.
However, the obtained atoms are not
guaranteed to be well localized in the graph since the learned  discrete values of $\widehat{g_s}$ are not necessarily derived from a
smooth kernel.  %Thus, the obtained dictionary is better in terms of approximation as it has more flexibility in adapting to the training signals. This result is particularly  obvious at low sparsity. However, as we relax the sparsity level we observe that the polynomial dictionary exhibits a better approximation performance.  
Moreover,  
the unstructured construction of the kernels in the method of   \cite{Zhang2012} leads to more complex implementations, as discussed in  Section \ref{practical_issues}.
%\ref{Computational_issues}. 

%\begin{figure*}[t]
%\begin{minipage}{2.1in}
%\begin{center}
%~ \includegraphics[width=1.1\textwidth]{Figures/coefficientsM400}~ \\
%~~~(a) M=400 ~\\
%\end{center}
%\end{minipage}
%\hfill
%\begin{minipage}{2.1in}
%\begin{center}
%~\includegraphics[width=1.1\textwidth]{Figures/coefficientsM600}~\\
%~~~(b) M=600 ~\\
%\end{center}
%\end{minipage}
%\hfill
%\begin{minipage}{2.1in}
%\begin{center}
%~\includegraphics[width=1.1\textwidth]{Figures/coefficientsM2000}~\\
%~~~(c) M=2000 ~\\
%\end{center}
%\end{minipage}
%\caption{Coefficients values of each subdictionary for different sizes of the training set. }  
%\label{fig:coeff}
%\vspace{-0.1cm}
%\end{figure*}

%\subsubsection{Polynomial degree and localization}
%Next, we examine the effect of the polynomial degree in the localization of the atoms on the graph.  In Fig. \ref{fig:coeff} we demonstrate the values of the polynomial coefficients of the learned kernels for the three different sizes of the training set of subsection \ref{TrainingSetExp1}. Note that in all the three cases the values of the coefficients for $k>5$ become significantly small. This is quite consistent with the maximum degree of $K=5$ imposed in the true dictionary. Thus, our algorithm is able to detect the localized patterns existing in the training signals even though, in the learning phase, we impose a degree that is much higher than the true degree of the polynomial.  tbc

%\subsubsection{Resilience to noise}

\begin{figure}
        \centering  
	   \subfigure[]{
	    \includegraphics[width=6.5cm]{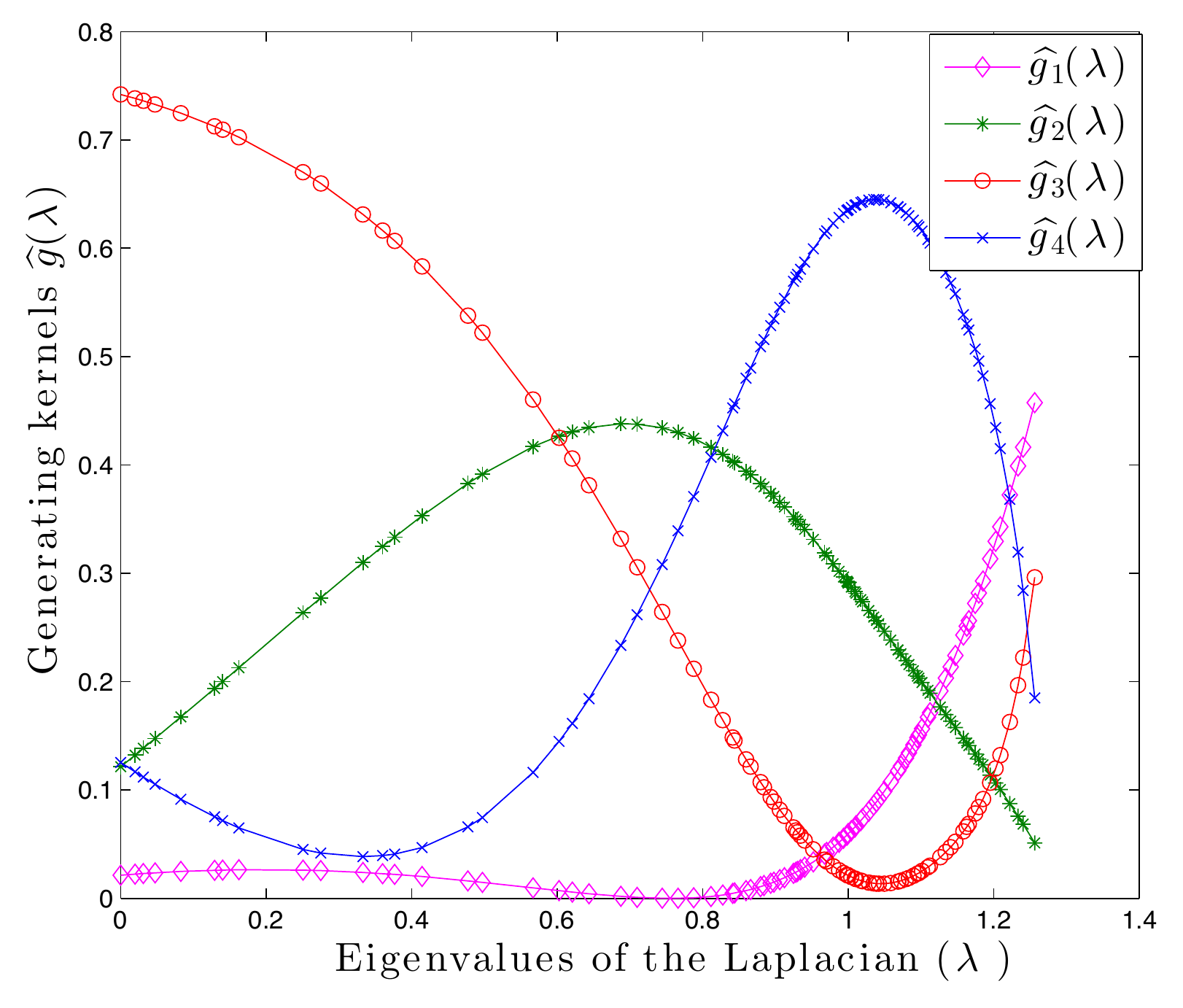}\label{fig:kernelsnoise0015}  
	    }	% \subfigure[Learned kernels]{ \includegraphics[width=7cm]{FiguresDictLear5/Synthetic1kernelsb.eps}\label{fig:kernelsl2b}}
         \subfigure[]{ \includegraphics[width=7cm]{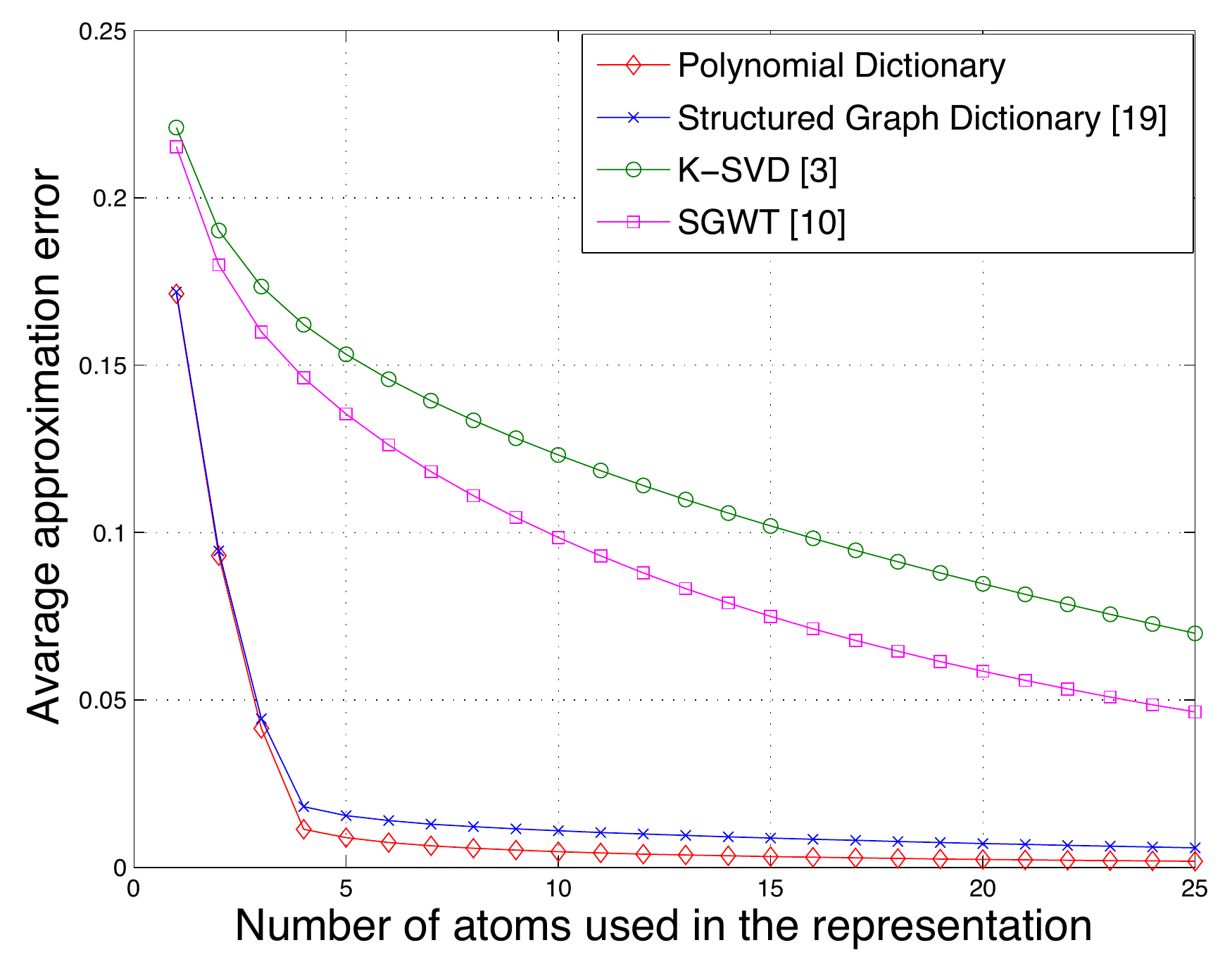}\label{fig:approxnoise0015}}
        \caption{Dictionary learning performance when  Gaussian noise  is added to the training signals.  (a) Kernels learned by the polynomial dictionary. (b) Approximation performance of the polynomial dictionary, SGWT\cite{Hammond2010}, K-SVD \cite{Aharon06} and the graph structured dictionary \cite{Zhang2012} on noiseless testing signals. } 
        \label{fig:approx_noisy}
\vspace{-0.5cm}
\end{figure}

Finally, we study the resilience of our algorithm to potential noise in the training phase. We generate $M=600$ training signals as linear combinations of $T_0\le 4$ atoms from the generating dictionary and add some Gaussian noise with zero mean and variance $\sigma=0.015$, which  corresponds to a noise level with $SNR=4dB$. These training signals are used to learn a polynomial dictionary with polynomials of degree $K=20$. The obtained kernels are shown in Fig. \ref{fig:kernelsnoise0015}. We observe that, even though the training signals are quite  noisy, we succeed in learning kernels that preserve the shape of our true noiseless kernels of Fig. \ref{original_kernels}.   We then test the performance of our algorithm on a set of 2000 testing signals generated as linear combinations of $T_0\le4$ atoms from the generating dictionary. The results for different sparsity levels in the OMP approximation are shown in Fig.  \ref{fig:approxnoise0015}. % compares the performance of our algorithm to that  obtained with SGWT, K-SVD and the structured graph dictionary of \cite{Zhang2012},  
%when a pre-defined number of atoms from the learned dictionary  is used in the approximation of the signals. 
%for different sparsity levels in the OMP approximation. 
We observe that both K-SVD and the structured graph dictionary are more sensitive to noise than our polynomial dictionary.

%\begin{figure}
%        \centering  
%	   \subfigure{
%	    \includegraphics[width=6.5cm]{Figures/kernelsM600K20noise01inkernel}\label{fig:kernelsnoise01}  
%	    }
%	    
%	         % \centering
%	          %\vspace{-.1in} 
%          {\small ~~~(a)}
%	
%	% \subfigure[Learned kernels]{ \includegraphics[width=7cm]{FiguresDictLear5/Synthetic1kernelsb.eps}\label{fig:kernelsl2b}}
%         \subfigure{ \includegraphics[width=6.5cm]{Figures/approximationM600K20noise01inkernel}\label{fig:approxnoise01}}
%        
%        	          %\centering
%	         % \vspace{-.1in} 
%          {\small ~~~(b)}
%        \caption{(a) Learned kernels with the polynomial dictionary and  (b) approximation performance of the polynomial dictionary, SGWT\cite{Hammond2010}, K-SVD \cite{Aharon06} and the graph structured dictionary \cite{Zhang2012}.  Gaussian noise of zero mean and $\sigma=0.1$ is added on the generating kernels  of the training signals while the testing signals are noiseless.} 
%        \label{fig:approx_noisy_in_kernels}
%\vspace{-0.5cm}
%\end{figure}

%\subsubsection{Influence of the translated atoms}

\subsubsection{Non-Polynomial Generating Dictionary}
In the next set of experiments, we depart from the idealistic scenario and study the performance of our polynomial dictionary learning algorithm in the more general case when the signal components are not   exactly polynomials of the Laplacian matrix. %We follow the same procedure as in the previous experiments in order to generate a graph of $N=100$ nodes. %   We generate a graph by randomly placing  $N=80$ vertices  in the unit square.   We design the edge weights   based on the thresholded Gaussian kernel function in such a way that $ W(i,j)=e^{-\frac{[dist(i,j)]^2}{2\theta^2}}$ if the physical distance between vertices $i$ and $j$ is less than or equal to $\kappa$, and zero otherwise.
%, \mbox { if }  dist(i,j) \le \kappa$, and 0 otherwise. 
%We fix $\theta=0.9$ and $\kappa=0.5$ in our experiments. %$w_{u,v}=e^{-\frac{[x_u-x_v]^2}{2\sigma^2}}$ if $ \|x_u-x_v\|\le \kappa$ and $w_{u,v}=0$ otherwise and we fix $\sigma=0.8$ and $\kappa=0.5$.  
%\begin{equation}
%{\small
%\notag W(u,v)= \left\lbrace
%\begin{array}{ll}
%		e^{-\frac{[x_u-x_v]^2}{2\sigma^2}}, & \mbox { if }  \|x_u-x_v\|\le \kappa
%\\0, & \mbox { otherwise }
%\end{array}
%\right.}
%\end{equation} 
%where we fix $\sigma=0.9$ and $\kappa=0.5$. 
%Moreover, 
In order to generate training and testing signals,
we divide the spectrum of the graph into four frequency bands, defined by the eigenvalues of the graph: $[\lambda_0:\lambda_{24}], [(\lambda_{25}:\lambda_{39})\cup (\lambda_{90}:\lambda_{99})],[\lambda_{40}:\lambda_{64}],$ and $[\lambda_{65}:\lambda_{89}]$. %,  each covering the eigenvalues $[1:20]$, $[21:30\quad 71:80]$, $[31:50]$, $[51:70]$ respectively.  
We then construct a generating  dictionary of $J=400$ atoms, with each atom having   a spectral representation that is  concentrated exclusively in one of the four bands. % covering one of the four bands of the dictionary. % by generating some random patterns, each of them covering one of the four bands of the spectrum. 
In particular, atom $j$ is of the form 
\begin{equation}
d_j=\widehat{h_j}(\mathcal{L})\delta_n=\chi \widehat{h_j}(\Lambda)\chi^{T} \delta_n.
\label{atom}
\end{equation}
 Each atom is generated independently of the others as follows. We randomly pick one of the four bands, randomly generate  25 coefficients uniformly distributed in the range $[0,1]$, and assign these random coefficients to be the diagonal entries of $\widehat{h_j}(\Lambda)$ corresponding to  the indices of the chosen spectral band. The rest of the values in  $\widehat{h_j}(\Lambda)$ are set to zero. The atom is then centered on a vertex $n$ that is also chosen randomly.  
Note that the obtained atoms are not guaranteed to be well localized in the vertex domain since the discrete values of $\widehat{h_j}(\Lambda)$ are chosen randomly and are not derived from a smooth kernel. Therefore, the atoms of the generating dictionary do not exactly match the signal model assumed by our dictionary design algorithm. 
% The resulting kernels are (i%)  not necessarily approximated by a polynomial function and (ii) not guaranteed to be smooth. They are thus not necessarily  well localized in the graph domain so that they do not exactly match the signal model considered in our dictionary design. % The kernel $\widehat{h_j}(\mathcal{L})$ is then placed in a random vertex $n$ on the graph i.e., $d_j=\widehat{h_j}(\mathcal{L})\delta_n$, generating the atom $j$ of the synthetic dictionary. % Finally, we place the  kernel in a random vertex $n$ on the graph, i.e., $d^{n}=\hat{g}_s(\L)\delta_n$. 
%
%Furthermore, each atom is placed in one random node of the graph. More precisely,  the atoms of our dictionary are of the  form $d^{n}=\hat{g}_s(\mathcal{L})\delta_n=\chi \hat{g}_s(\Lambda)\chi^{T} \delta_n$, where $\hat{g}_s$ is the kernel covering the spectrum of the $s^{th}$ band and $n$ is the node where the atom is placed.    %Thus, in order to generate a  kernels $g_j(\mathcal{L})$, we need to randomly generate $g_j(\Lambda)$.  
%Each atom  is generated in the following way. First, we pick   $s$ randomly, and  we generate $20$ uniformly random coefficients that cover the diagonal entries of $\hat{g}_s(\Lambda)$ corresponding to  $i$. %which means that this kernel will cover only the spectrum between $[20(j-1)+1:20j]$. We choose randomly $5$ positions in the range $[20(j-1)+1:20j]$ to which we attribute  uniformly random coefficients in $[0:1]$. 
%The rest of the entries  are set to zero. Finally, we place the  kernel in a random vertex $n$ on the graph, i.e., $d^{n}=\hat{g}_s(\L)\delta_n$.  
Finally, we generate the  training  signals by linearly combining (with random coefficients) $T_0\le4$ random atoms from the generating dictionary. %In the following, we examine the influence of two different parameters in the performance of our learning algorithm: the maximum degree of the polynomial and the size of the training set.  %, with one atom from each band in each of the training signals. %,  and we add some additive Gaussian noise of zero mean and $\sigma=0.05$. 
%\subsubsection{Effect of the polynomial degree}

\begin{figure}
      \centering
	   \subfigure[$K=5$]{ \includegraphics[width=7cm]{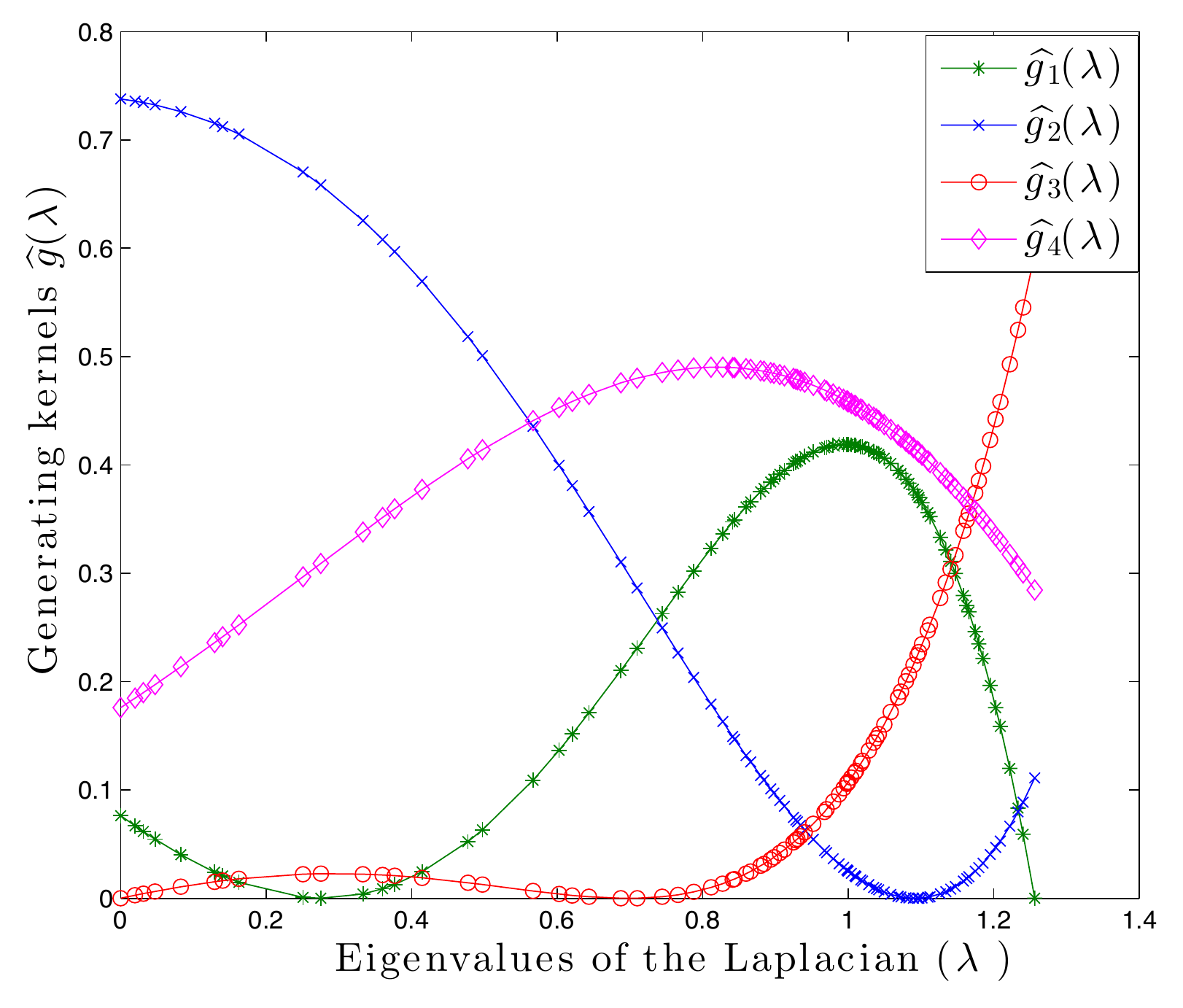}\label{fig:K5}}
	 \subfigure[$K=10$]{ \includegraphics[width=7cm]{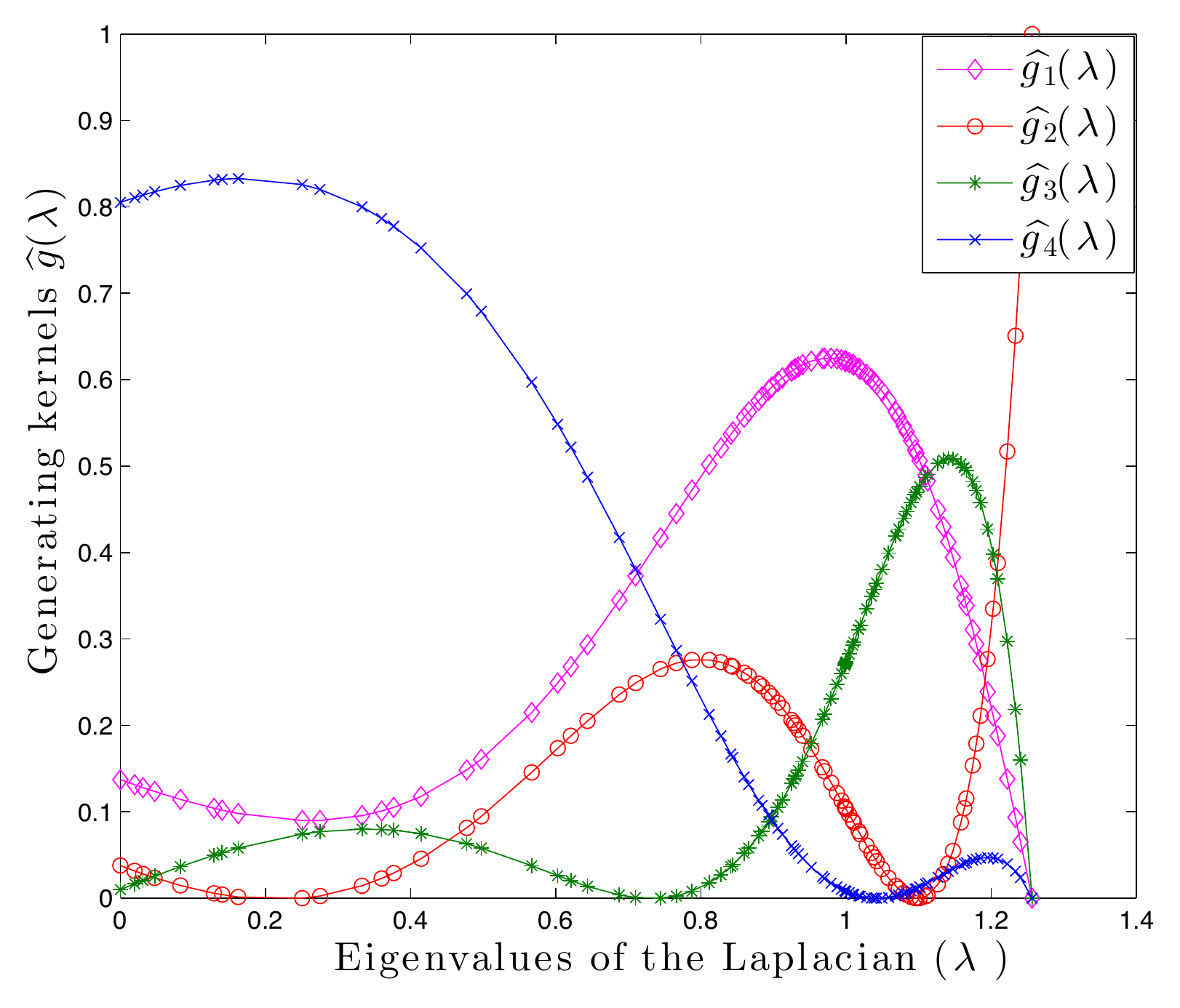}\label{fig:K10}}
          \subfigure[$K=20$]{ \includegraphics[width=7cm]{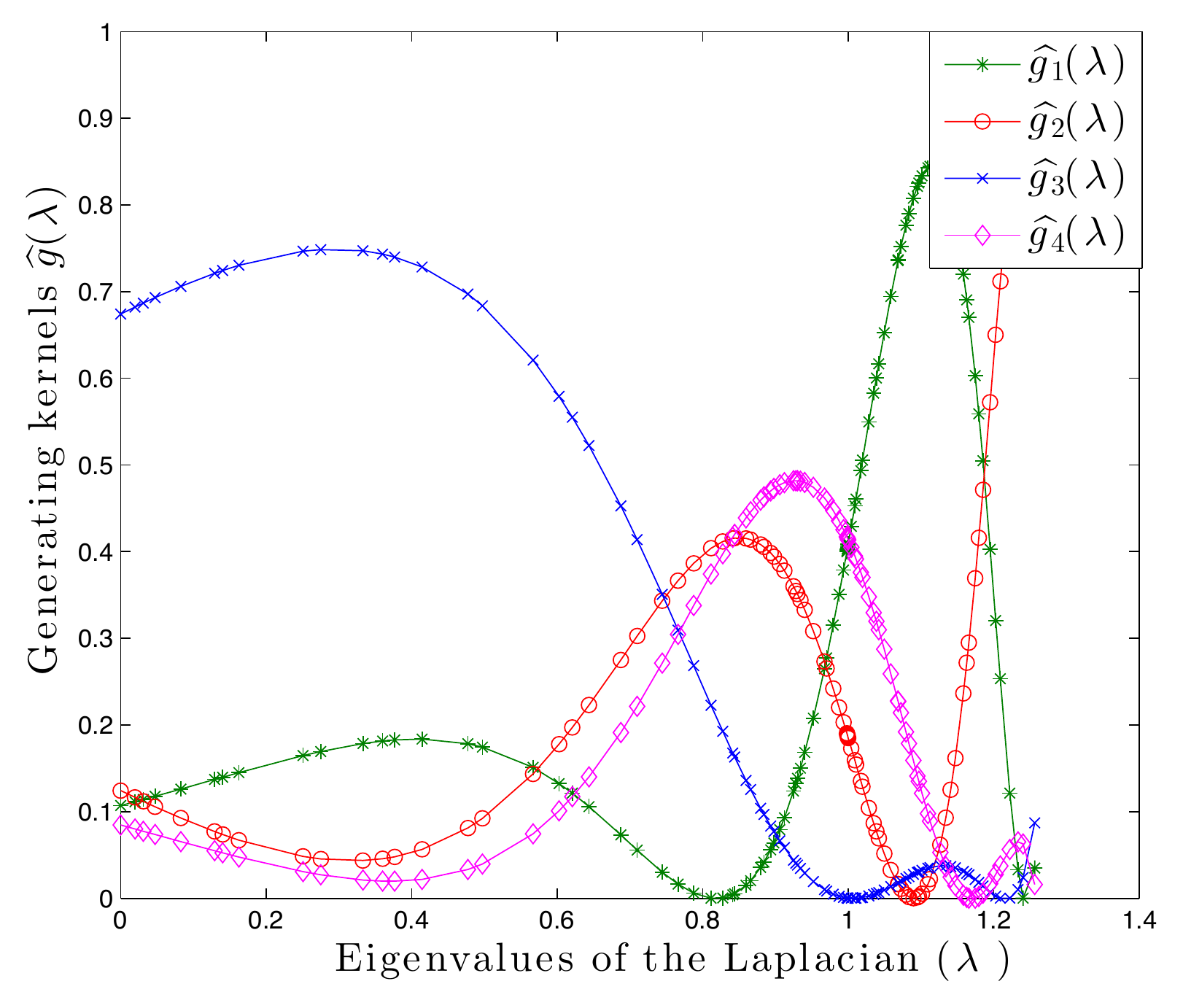}\label{fig:K20}}
           \subfigure[$K=25$]{ \includegraphics[width=7cm]{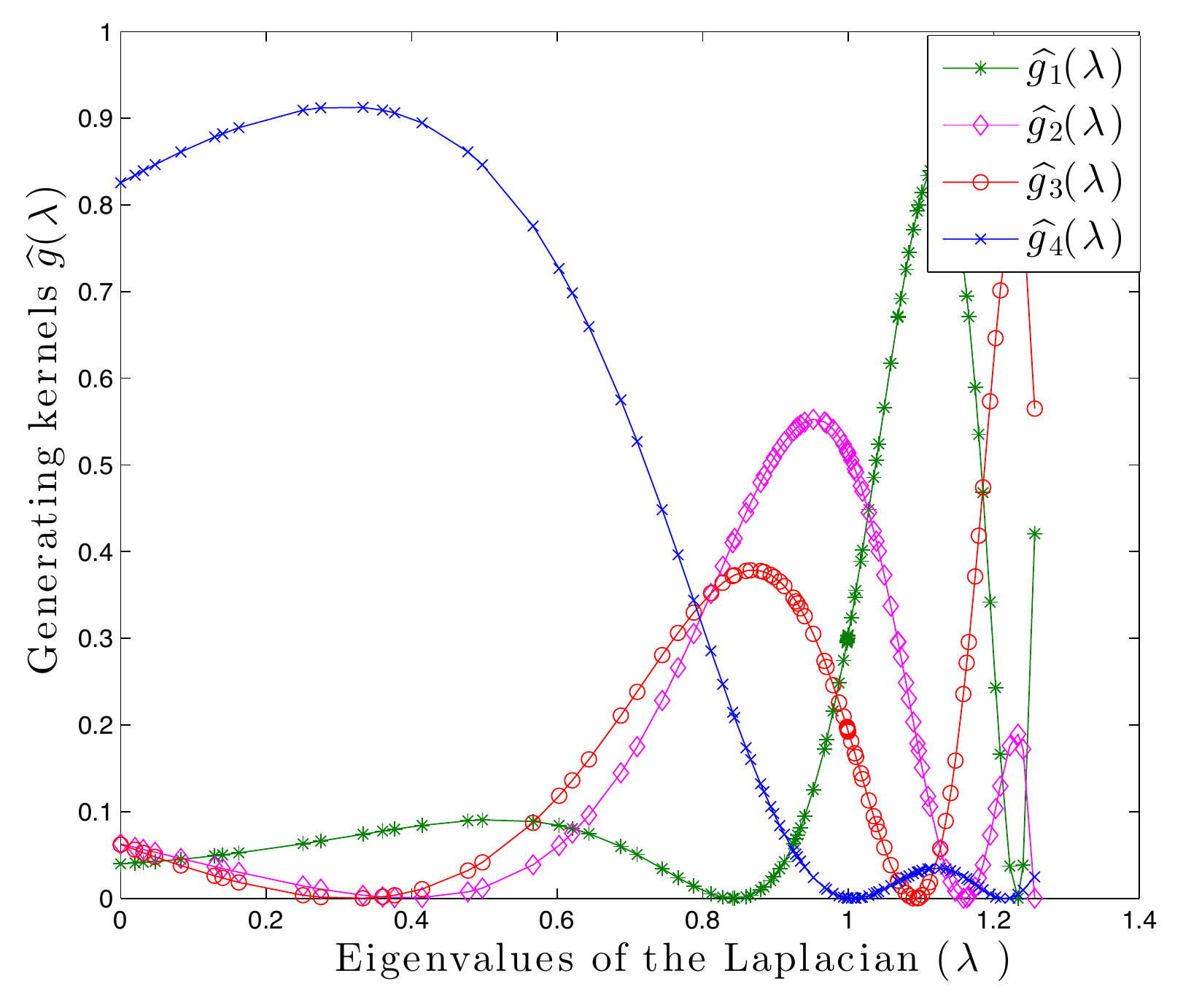}\label{fig:K25}}
        \caption{Kernels $\left\{\widehat{g_s}(\cdot)\right\}_{s=1,2,3,4}$ learned by the polynomial dictionary algorithm for (a) $K=5$, (b) $K=10$, (c) $K=20$ and (d) $K=25$. } 
        \label{fig:kernels_diff_degree}
\vspace{-0.5cm}
\end{figure}

%In these non-ideal settings, we first study the effect of the polynomial degree on (i)   the learned kernels and (ii) the representation of graph signals. 
%In particular, w
We first verify the ability of our dictionary learning algorithm to recover the spectral bands that are used  in the synthetic generating dictionary.   %want to verify that our dictionary learning algorithm is able to recover the spectral bands imposed in the synthetic dictionary. In order to do so, we examine the  relation between the obtained kernels and the  maximum degree $K$ of the polynomials. 
We fix the number of training signals to $M=600$ and run our dictionary learning algorithm for four different degree values  of the polynomial, i.e., $K=\{5,10,20, 25\}$. The kernels $\left\{\widehat{g_s}(\cdot)\right\}_{s=1,2,3,4}$ obtained  for the four subdictionaries are shown in Fig. \ref{fig:kernels_diff_degree}. We observe that for higher values of $K$, the learned kernels are more localized in the graph spectral domain.
%the selectivity of the kernels in the spectral domain depends on the degree of the polynomial; in particular, 
%the higher the degree, the more localized  the spectral representation of the kernels.   
We further notice that,  for $K=\{20, 25\}$, each  kernel approximates  one of the four bands defined in the generating dictionary, similarly to the behavior of classical frequency filters. %a  filter behavior. 
%This result is not surprising:  by %relaxing the localization in the graph domain, i.e., by
% increasing the degree of the polynomials, we give more flexibility to our learning algorithm to select the correct kernels.   Our algorithm is thus able to recover the spectral components that dominate the training signals   by learning  kernels that are selective in the spectral domain. 
 
\begin{figure}
      \centering
	   \subfigure[$\widehat{g_{1}}(\L)\delta_{1}$]{ \includegraphics[width=7cm]{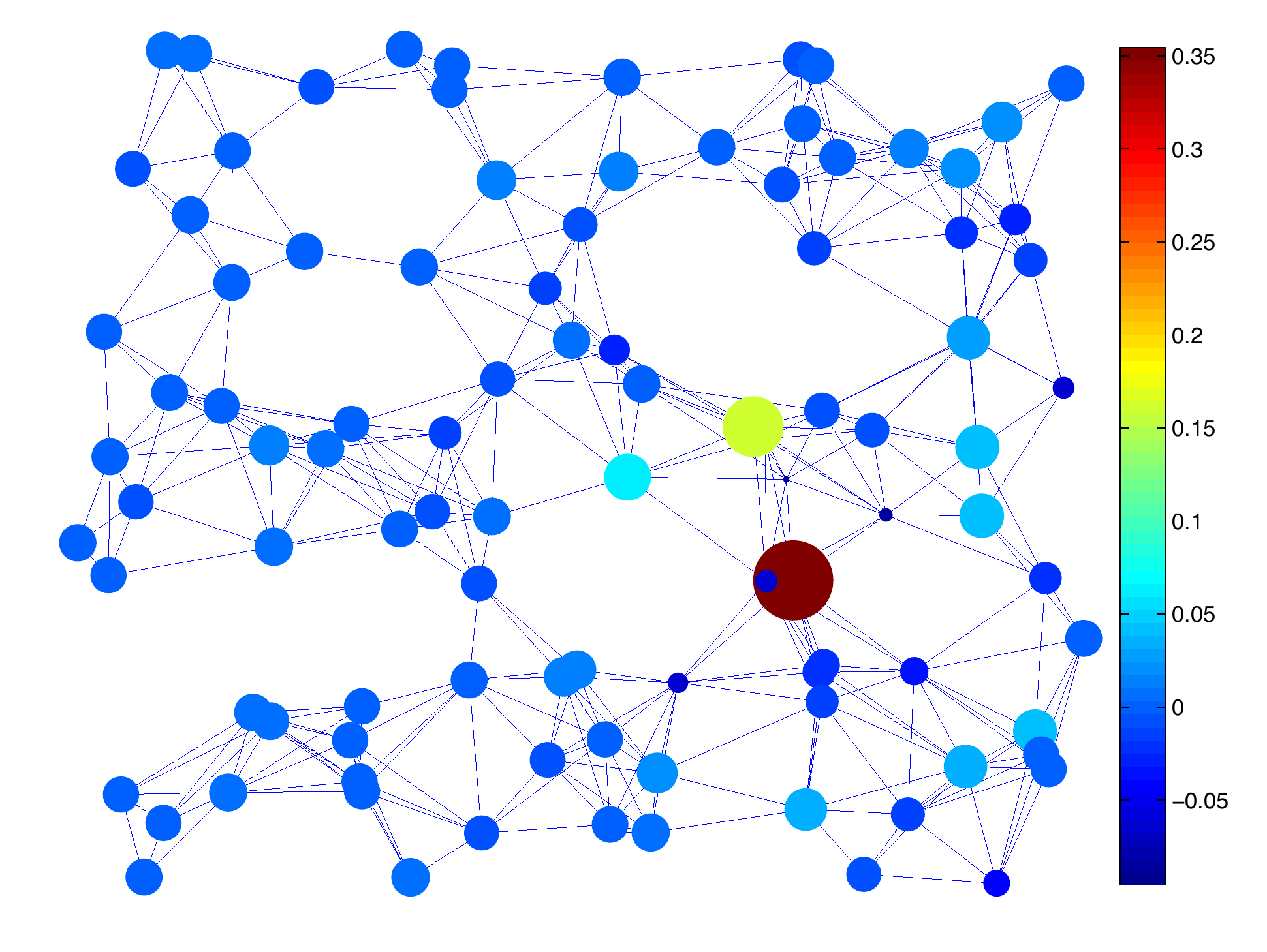}\label{fig:atom1}}
	 \subfigure[$\widehat{g_{2}}(\L)\delta_{1}$]{ \includegraphics[width=7cm]{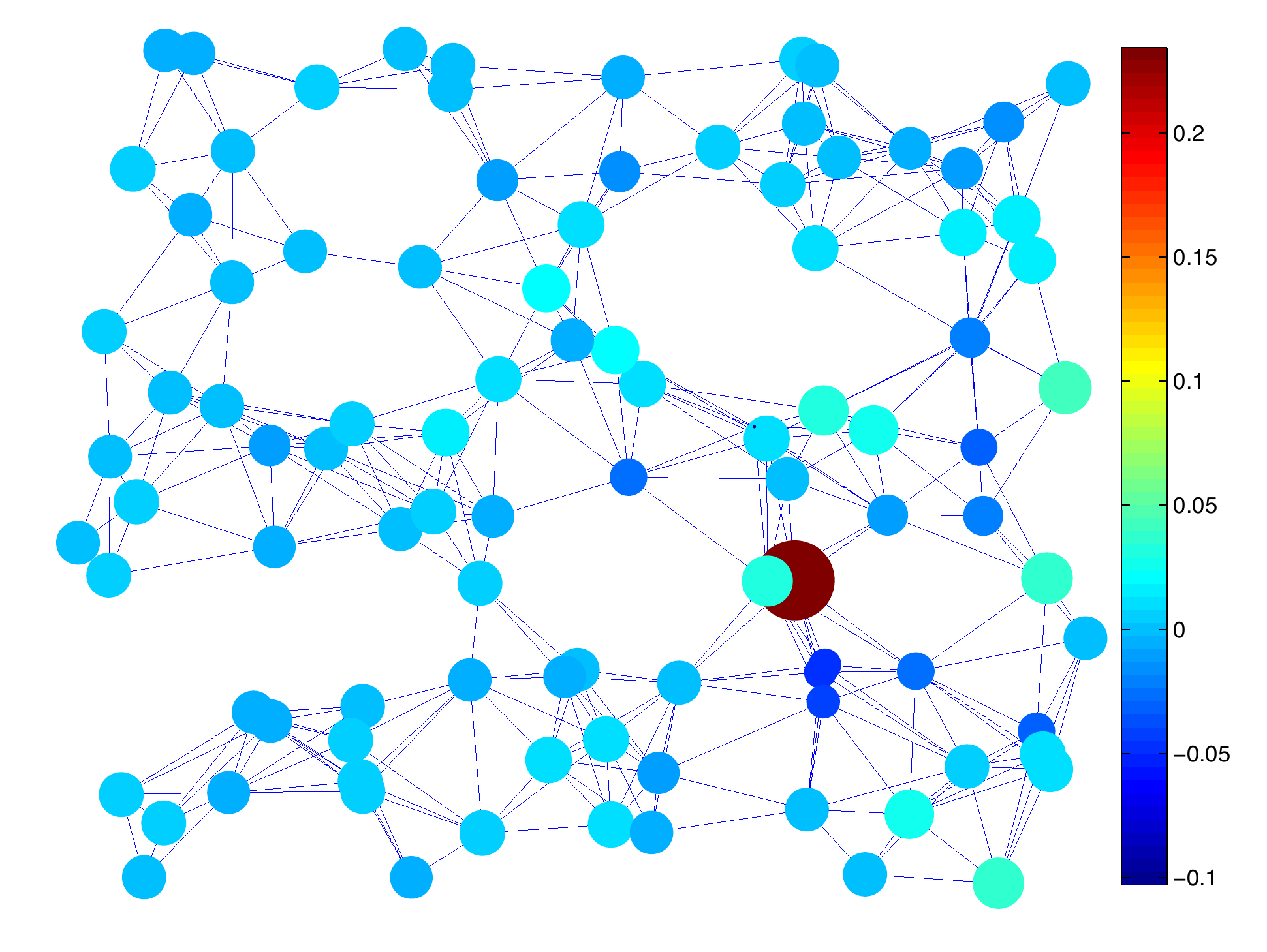}\label{fig:atom2}}
          \subfigure[$\widehat{g_{3}}(\L)\delta_{1}$]{ \includegraphics[width=7cm]{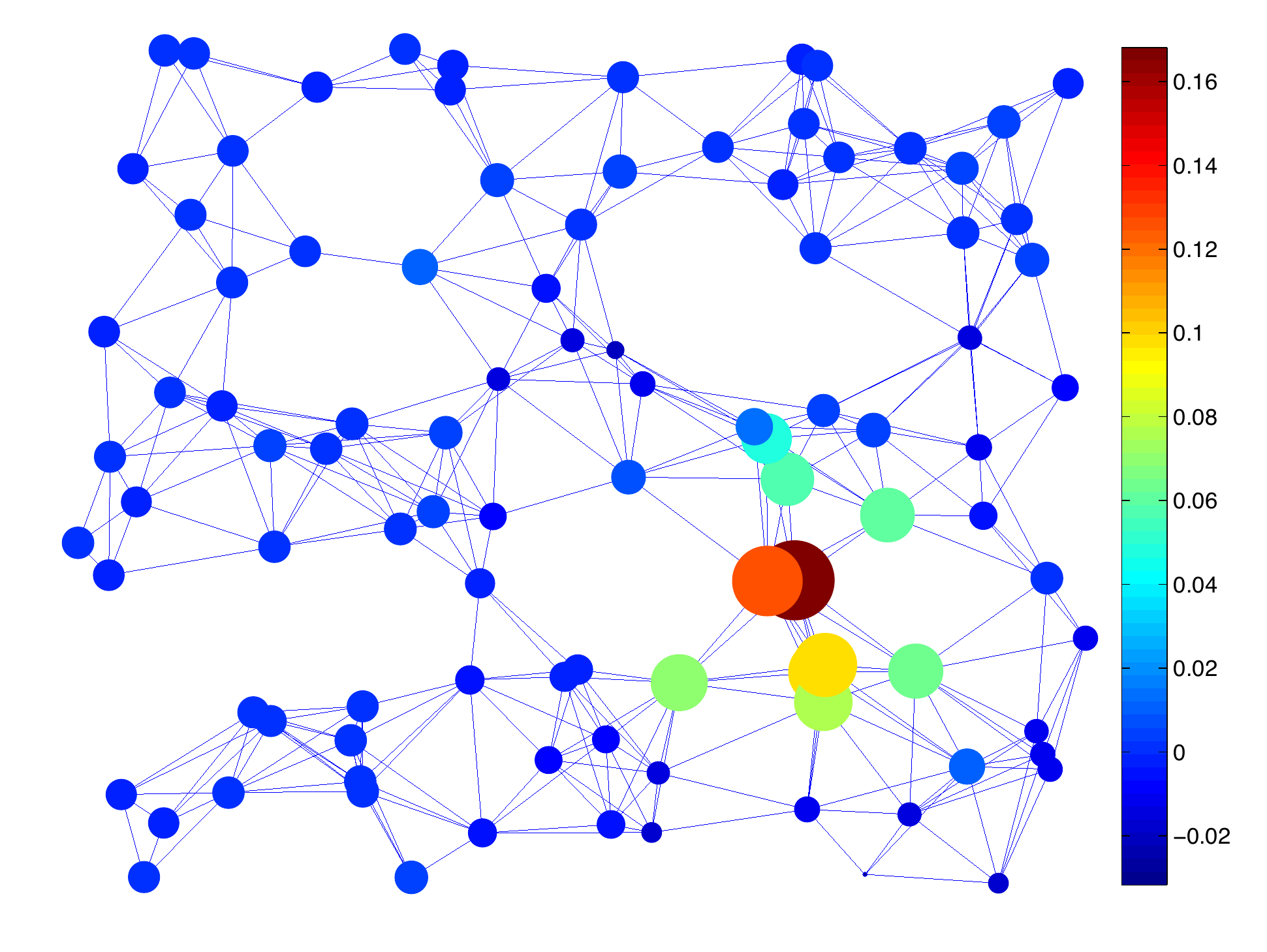}\label{fig:atom3}}
           \subfigure[$\widehat{g_{4}}(\L)\delta_{1}$]{ \includegraphics[width=7cm]{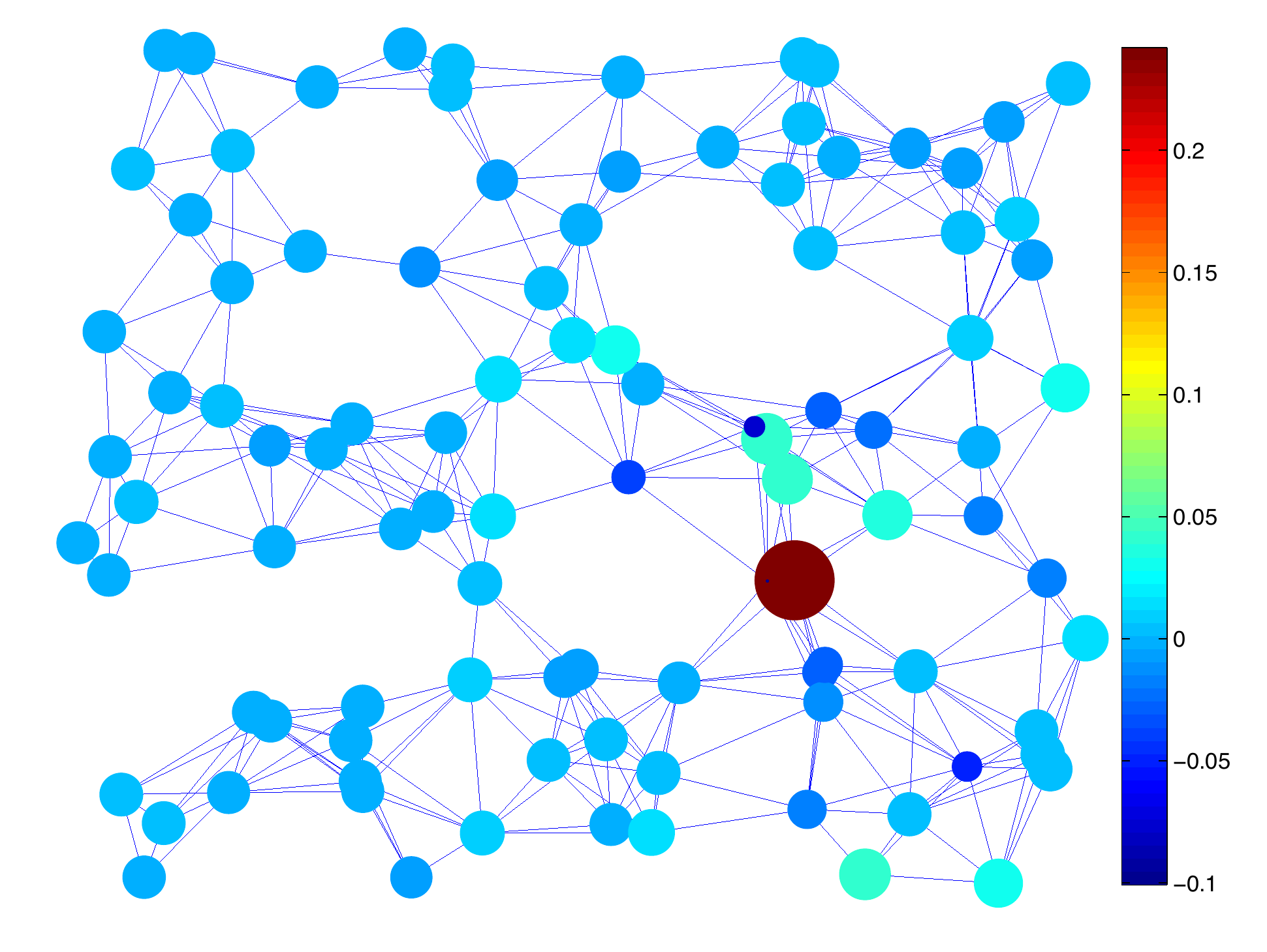}\label{fig:atom4}}
        \caption{Learned atoms centered on vertex $n=1$, from each of  the subdictionaries. The size and the color of each ball indicate the value of the atom in each vertex of the graph.  } 
        \label{fig:atoms}
\vspace{-0.5cm}
\end{figure}

In Fig. \ref{fig:atoms},  we illustrate  the four learned atoms centered at the vertex $n=1$ (one atom for each subdictionary), with  $K=20$.  
%obtained in each of the four subdictionaries for a polynomial kernel of  degree $K=20$ and positioned  at the location $n=1$. 
We can see that the support of the atoms adapts to the graph topology. %and diffuse locally around one particular vertex. 
The atoms can be either smoother around a particular vertex, as for example in Fig. \ref{fig:atom3}, or more localized, as in Fig. \ref{fig:atom1}. 
Comparing Figs. \ref{fig:K20}, and \ref{fig:atoms}, we observe that  the localization of the atoms in the graph domain depends on the spectral behavior of  the kernels.  
Note that the smoothest atom on the graph (Fig. \ref{fig:atom3}) corresponds to the subdictionary generated from the kernel that is concentrated on the low frequencies (i.e., $\widehat{g_3}(\cdot)$). This is because
%result derives from  the fact that 
the graph Laplacian eigenvectors associated with the lower frequencies are smoother with respect to the underlying graph topology, while those associated with the larger eigenvalues oscillate more rapidly \cite{Shuman13}.  % while the most localized atom (Fig. \ref{fig:atom3}) $. 

%We can thus conclude that  there is a tradeoff between the localization of the atoms in the vertex domain and the selectivity of the kernels in the spectral domain. 
 
Next, we test the approximation performance of our learned dictionary on a set of 2000 testing signals generated  in exactly the same way as the training signals.    %The average approximation error is set to $\|Y_{test}-\D X\|_F^2/|Y_{test}|$, where $|Y_{test}|$ is the size of the testing set and is measured for a fixed sparsity level.   
    Fig. \ref{fig:approximation_different_degrees} shows that the approximation performance obtained with our algorithm improves as we increase the polynomial degree. This is attributed to two main reasons: (i) by increasing the polynomial degree, we allow more flexibility in the learning process; (ii)  a small $K$ implies that  the atoms are localized in a small neighborhood and thus more atoms are needed to represent signals with support in different areas of the graph.
%    {\color{red}[But you could also change the sparsity level of OMP accordingly, right? I assume this is for a fixed sparsity level?]}
     %cover the entire graph. %An example of the localization properties of the obtained atoms  is shown in  Fig. \ref{fig:atoms}.  
    
 \begin{figure}
      \centering
          { \includegraphics[width=7cm]{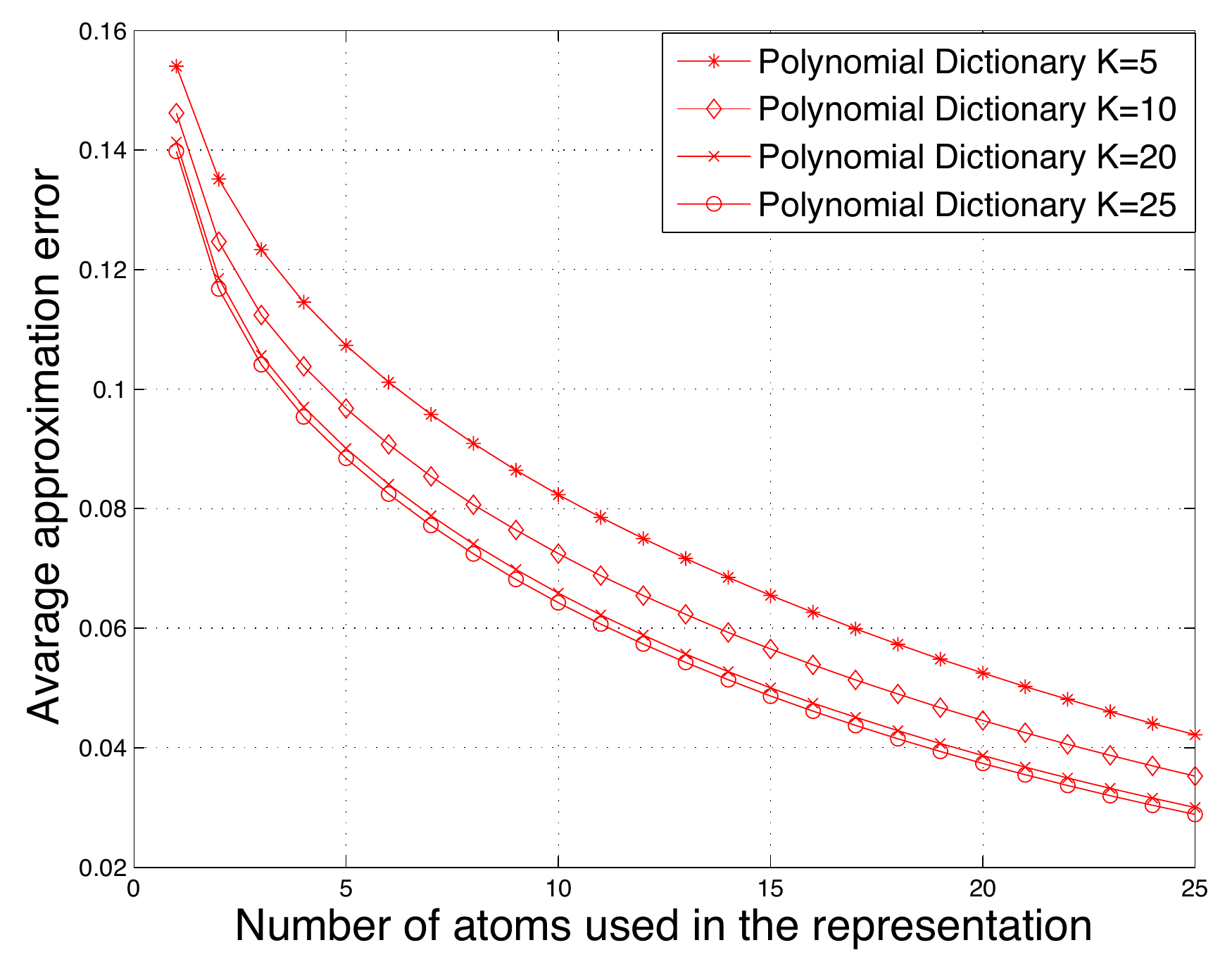}\label{fig:approximation_different_degrees}}
        \caption{Comparison of the average approximation performance of our learned dictionary on test signals generated by the non-polynomial synthetic generating dictionary, for $K=\{5,10,20, 25\}$.} 
        \label{fig:approximation_different_degrees}
%\vspace{-0.5cm}
\end{figure}

\begin{figure*}[t]
\begin{minipage}{2.1in}
\begin{center}
~ \includegraphics[width=1.05\textwidth]{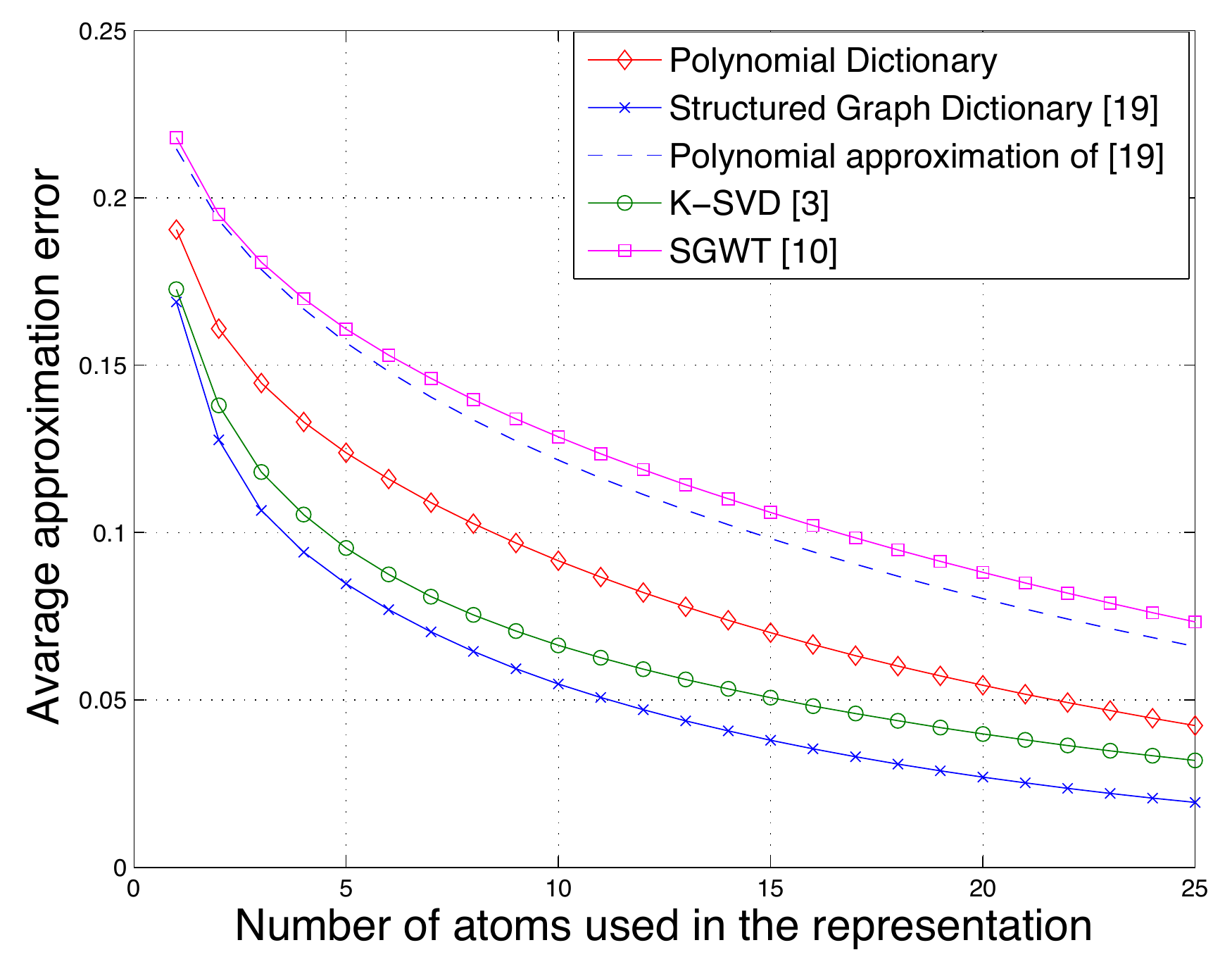}~ \\
~~~(a) M=400 ~\\
\end{center}
\end{minipage}
\hfill
\begin{minipage}{2.1in}
\begin{center}
~\includegraphics[width=1.05\textwidth]{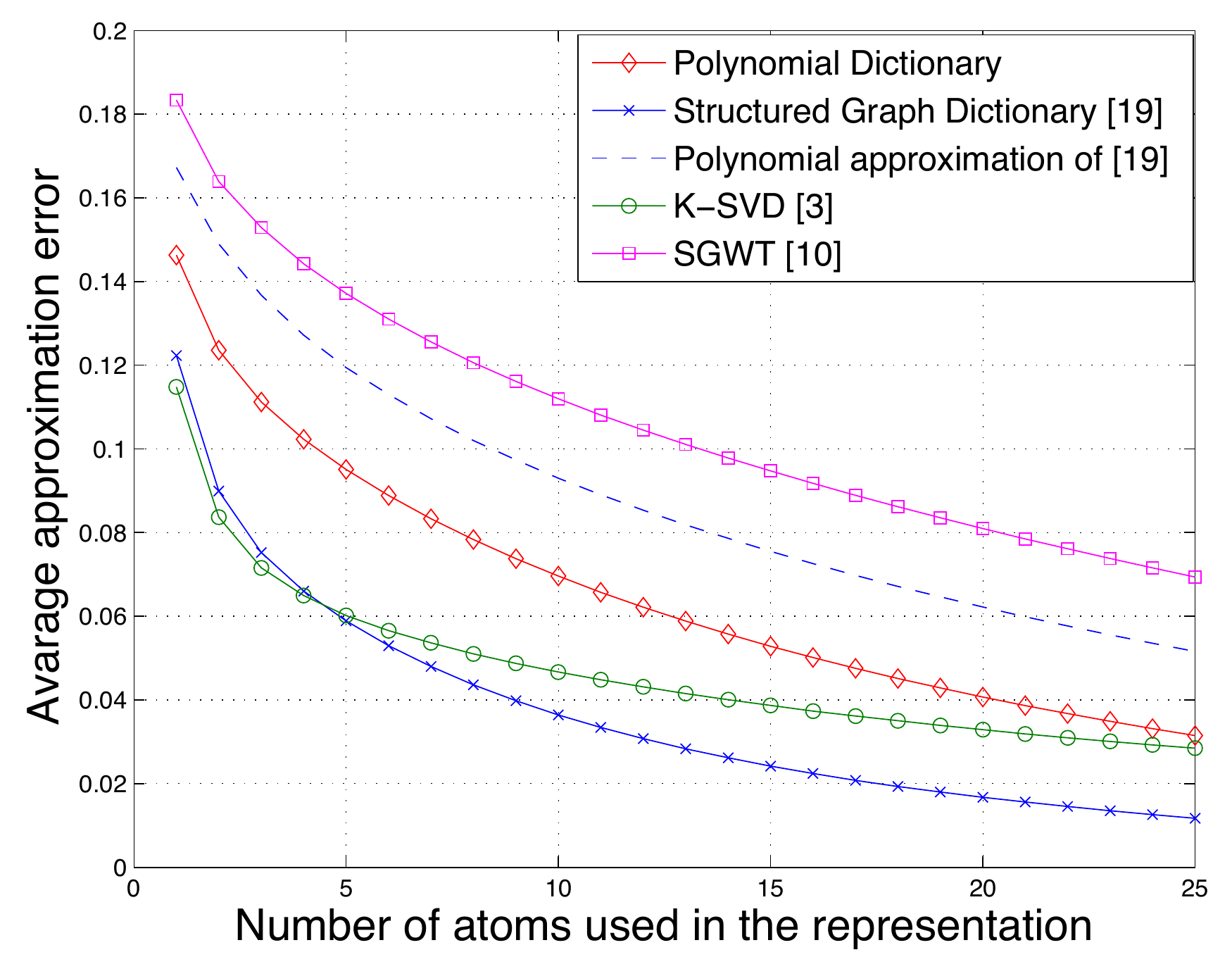}~\\
~~~(b) M=600 ~\\
\end{center}
\end{minipage}
\hfill
\begin{minipage}{2.1in}
\begin{center}
~\includegraphics[width=1.05\textwidth]{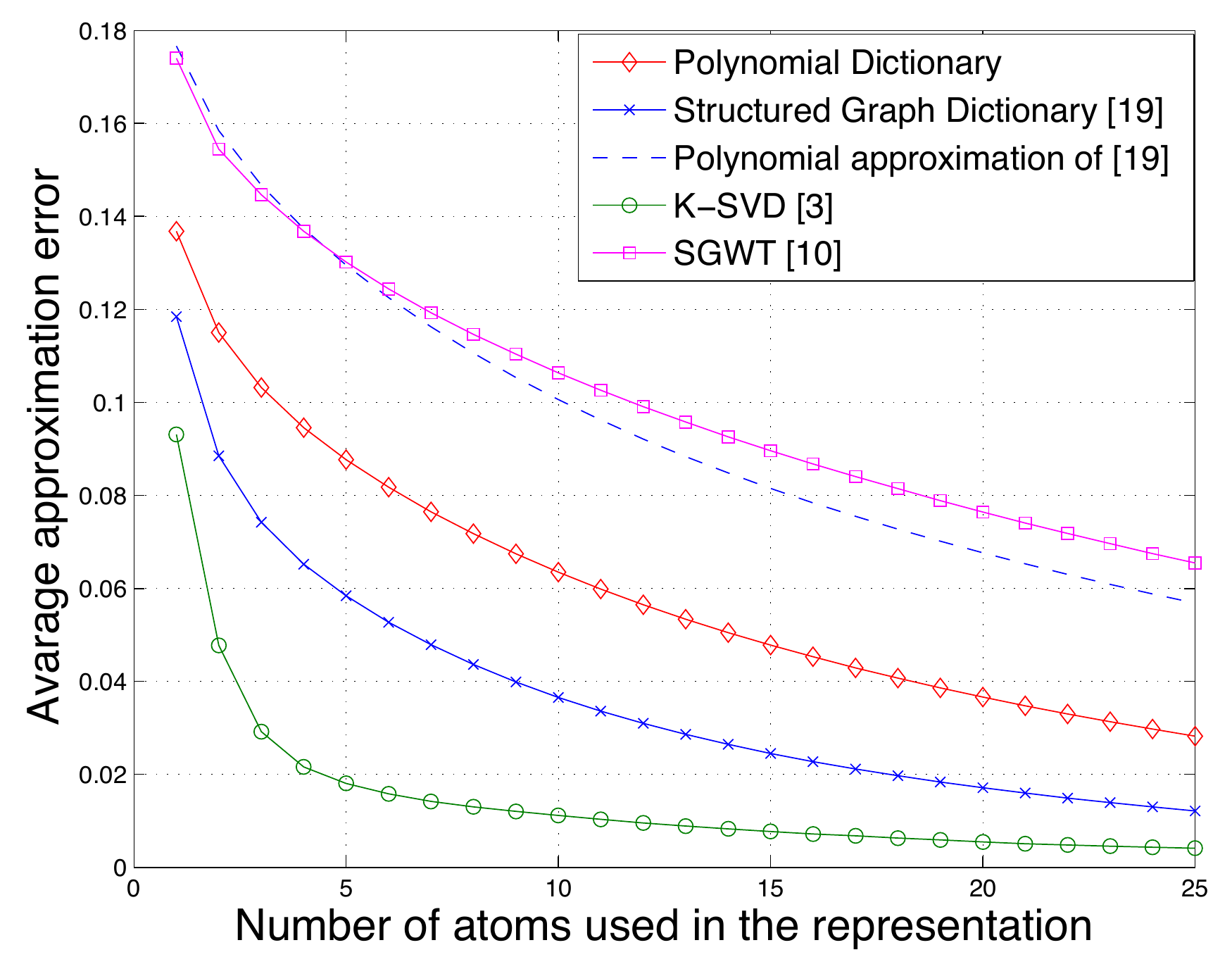}~\\
~~~(c) M=2000 ~\\
\end{center}
\end{minipage}
\caption{Comparison of the learned polynomial dictionary to the SGWT\cite{Hammond2010}, K-SVD \cite{Aharon06} and the graph structured dictionary \cite{Zhang2012} and its polynomial approximation in terms of approximation performance on test data generated from a non-polynomial generating dictionary, for different sizes of the training set. }  
\label{fig:difM}
\vspace{-0.1cm}
\end{figure*}

In Fig. \ref{fig:difM}, we fix $K=20$, and compare the approximation performance of our learned dictionary to that of other dictionaries, with exactly the same setup as we used in Figure \ref{fig:difM1}. We again observe that K-SVD is the most sensitive to the size of the training data, for the same reasons explained earlier. Since the kernels used in the generating dictionary in this case do not match our polynomial model, the structured graph dictionary learning algorithm of \cite{Zhang2012} has more flexibility to learn the non-smooth generating kernels and therefore generally achieves better approximation. Nonetheless, the approximation performance of our learned dictionary is competitive, especially for smaller training sets, and the learned dictionary is more computationally efficient to implement.  
For a fairer comparison of approximation performance, we fit an order $K=20$ polynomial function to the discrete values $\widehat{g_s}$ learned with the algorithm of \cite{Zhang2012}. We observe that our polynomial dictionary outperforms the polynomial approximation of the dictionary learned by \cite{Zhang2012} in terms of approximation performance.

\subsubsection{Generating Dictionary Focused on Specific Frequency Bands}
\begin{figure}
      \centering
	  { \includegraphics[width=7cm]{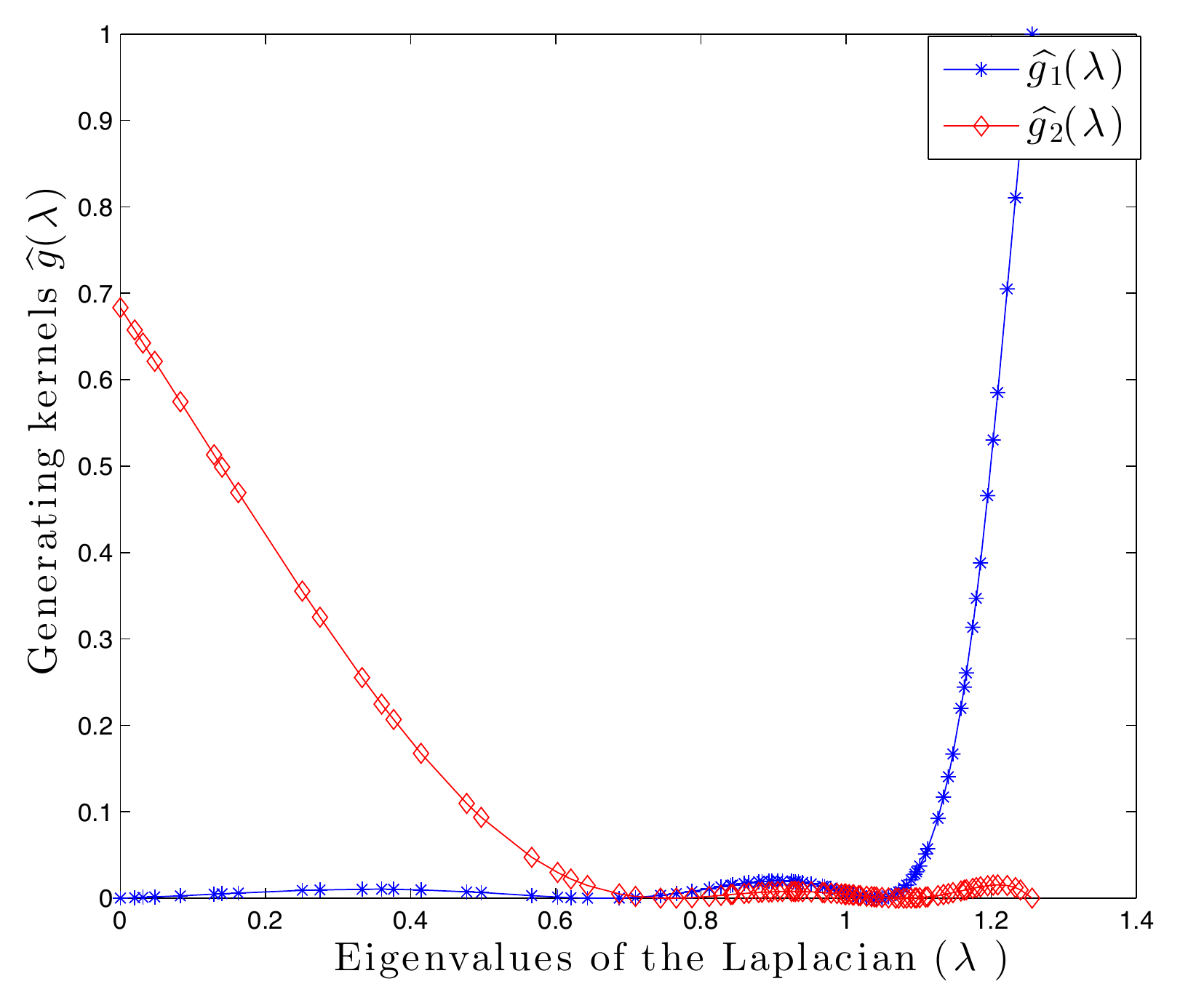}\label{fig:kernels_2bands1}}
	% \subfigure[$K=10$]{ \includegraphics[width=7cm]{Figures/kernels_2bands}\label{fig:kernels_2bands2}}
        \caption{Kernels $\left\{\widehat{g_s}(\cdot)\right\}_{s=1,2}$ learned by the polynomial dictionary algorithm  from a set of training signals that are supported in only two particular bands of the spectrum. The bands are defined by the eigenvalues of the graph $[\lambda_0:\lambda_{9}]$ and $[\lambda_{89}:\lambda_{99}]$, which correspond to the values  $[0: 0.275]$ and $[1.174: 1.256]$ respectively. } 
        \label{fig:kernels_2bands}
\vspace{-0.5cm}
\end{figure}

In the final set of experiments, we study the behavior of our algorithm in the case when we have the additional prior information that the training signals do not cover the entire spectrum, but are concentrated only in some bands that are not known \emph{a priori}.  In order to generate the training signals,
we choose only two particular  frequency bands, defined by the eigenvalues of the graph: $[\lambda_0:\lambda_{9}]$ and $[\lambda_{89}:\lambda_{99}]$, which correspond to the values  $[0: 0.275]$ and $[1.174: 1.256]$, respectively.  We construct a generating  dictionary of $J=400$ atoms, with each atom concentrated in only one of the two bands and generated according to (\ref{atom}). The training signals ($M=600$)  are then constructed by linearly combining $T_0\le4$ atoms from the generating dictionary. We set $K=20$ and $\epsilon_1=c$ in order to allow our polynomial dictionary learning algorithm the flexibility to learn kernels 
%We use our polynomial dictionary learning algorithm to learn a dictionary of $K=20$ for sparsely representing the training signals and we set $\epsilon_1=c$ in order to allow the flexibility to our algorithm to learn kernels 
that are supported only on specific frequency bands. The learned kernels are illustrated in Fig. \ref{fig:kernels_2bands}. We observe that the algorithm is able to detect the spectral components that exist in the training signals since the learned kernels are concentrated only in the two parts of the spectrum to which the atoms of the generating dictionary belong.  

\subsection{Approximation of Real Graph Signals}\label{realworld}
After examining the behavior of the polynomial dictionary learning algorithm for  synthetic signals,  we illustrate the performance of our algorithm  in the approximation of localized graph signals from  real world datasets. In particular, we examine the following three datasets.  
\begin{itemize}
\item {\textbf{Flickr Dataset:} }
 We consider the daily number of distinct Flickr users that took photos %for every day  
 at different geographical locations around  Trafalgar Square in  London,  between January 2010 and June 2012  \cite{Dong2013}. Each vertex of the graph represents a geographical  area of $10\times10$ meters. The graph is constructed by assigning an edge between two locations when the distance between them is shorter than 30 meters, and the edge weight is set to be inversely proportional to the distance.  We then remove edges whose weight is below a threshold in order to obtain a sparse graph. The number of vertices of the graph is $N=245$. We set  $S=2$ and  $K=10$ in our dictionary learning algorithm. We have a total of 913 signals, and we use  700 of them for training and the rest for testing.    % We test the approximation performance of our dictionary in the testing signals and we compare it  to that obtained with (i) graph based transform methods such as the spectral graph wavelet transform \cite{Hammond2010} , (ii) classical dictionary learning methods such as K-SVD \cite{Aharon06} and (iii)  the graph based dictionary learning algorithm presented in \cite{Zhang2012}, for an increasing sparsity level. In order to be consistent with both algorithms  in \cite{Aharon06} and   \cite{Zhang2012}, the sparse coding step is performed using OMP. 
 
\item {\textbf{Traffic Dataset:}}
We consider the daily bottlenecks in Alameda County in California between January 2007 and May 2013. The data are part of the Caltrans Performance Measurement System (PeMS) dataset that provides traffic information throughout all major metropolitan areas of California \cite{Choe2002}.\footnote{The data are  publicly available at http://pems.dot.ca.gov.} In particular, the nodes of the graph consist of $N=439$ detector stations where bottlenecks were identified over the  period under consideration. The graph is designed by connecting stations when the distance between them is smaller than a threshold of $\theta=0.08$ which corresponds to approximately 13 kilometres. The distance is set to be the Euclidean distance of the GPS coordinates of the stations and the edge weights are set to be inversely proportional to the distance. A bottleneck could be any location where there is a persistent drop in speed, such as  merges,  large on-ramps, and incidents. The signal on the graph is the average length of the time in minutes that a bottleneck is active for each specific day.   In our experiments, we fix the maximum degree of the polynomial to $K=10$ and we learn a dictionary consisting of $S=2$ subdictionaries. We use the signals in the period between January 2007 and December 2010 for training and the rest for testing. For computational issues, we normalize all the signals with respect to the norm of the  signal with maximum energy. 

\item {\textbf{Brain Dataset:}} We consider a set of fMRI signals acquired on five different subjects \cite{EryilmazVSV11}, \cite{RichiardiESVV11}. For each subject, the signals have been preprocessed into timecourses of $N=90$ brain regions of contiguous voxels, which are determined from a fixed anatomical atlas, as described in  \cite{RichiardiESVV11}.  The timecourses for each subject correspond to 1290 different graph signals that are measured while the subject is in different  states, such as completely relaxing, in the absence of any stimulation or  passively watching small movie excerpts. 
%Depending on the state of each subject, there are some events happening over time and in different parts of the brain.  
For the purpose of this paper, %{\color{red}[Perhaps add a footnote saying that it would be more interesting to consider the entire dataset as a 3D dataset, and try to learn a dictionary that accounts for both temporal correlations and correlations across different brains regions according to the graph structure.]}, 
we treat the measurements at each time as an independent signal on the 90 vertices of the brain graph. The anatomical distances between regions of the brain are approximated by the Euclidean distance between the coordinates of the centroids of each region, the connectivity of the graph is determined  by assigning an edge between two regions when the anatomical distance between them is shorter than 40 millimetres, and the edge weight is set to be inversely proportional to the distance.
We then apply our polynomial dictionary learning algorithm in order to learn a dictionary 
of atoms representing brain activity across the network at a fixed point in time.
%that is able to capture the events happening in the fMRI signals. 
%In particular, we design the brain graph based on the anatomical distances between the brain regions. The anatomical distances are approximated by the Euclidean distance between the coordinates of the centroids of each region.  The connectivity of the graph is determined  by assigning an edge between two regions when the anatomical distance between them is shorter than 40, and the edge weight is set to be inversely proportional to the distance. 
We use the graph signals from the timecourses of two subjects as our training signals and we learn a dictionary of $S=2$ subdictionaries and a maximum polynomial degree of $K=15$.  We use the graph signals from the remaining three timecourses to  validate the performance of the learned dictionary. As in the previous dataset, we normalize all of the graph signals with respect to the norm of the  signal with maximum energy. 
 
 \end{itemize} 
 
% \begin{figure*}[t]
%\begin{minipage}{2.1in}
%\begin{center}
%~ \includegraphics[width=1.05\textwidth]{Figures/flickr_example1_reduced}~ \\
%~~~(a) Flickr Dataset ~\\
%\end{center}
%\end{minipage}
%\hfill
%\begin{minipage}{2.1in}
%\begin{center}
%~\includegraphics[width=1.05\textwidth]{Figures/traffic_example_reduced}~\\
%~~~(b) Traffic Dataset ~\\
%\end{center}
%\end{minipage}
%\hfill
%\begin{minipage}{2.1in}
%\begin{center}
%~\includegraphics[width=1.05\textwidth]{Figures/brain_example_reduced}~\\
%~~~(c) Brain Dataset ~\\
%\end{center}
%\end{minipage}
%\caption{Examples of  graph signals from three different datasets: (a) Number of distinct Flickr users that have taken photos near Trafalgar square in one day. (b) Daily bottleneck (in minutes) in Alameda County. (c) An fMRI signal acquired from a subject.  }  
%\label{fig:dataset_examples}
%\vspace{-0.1cm}
%\end{figure*}
%
 
\begin{figure}
      \centering
         \subfigure [Flickr Dataset] {\includegraphics[width=7cm]{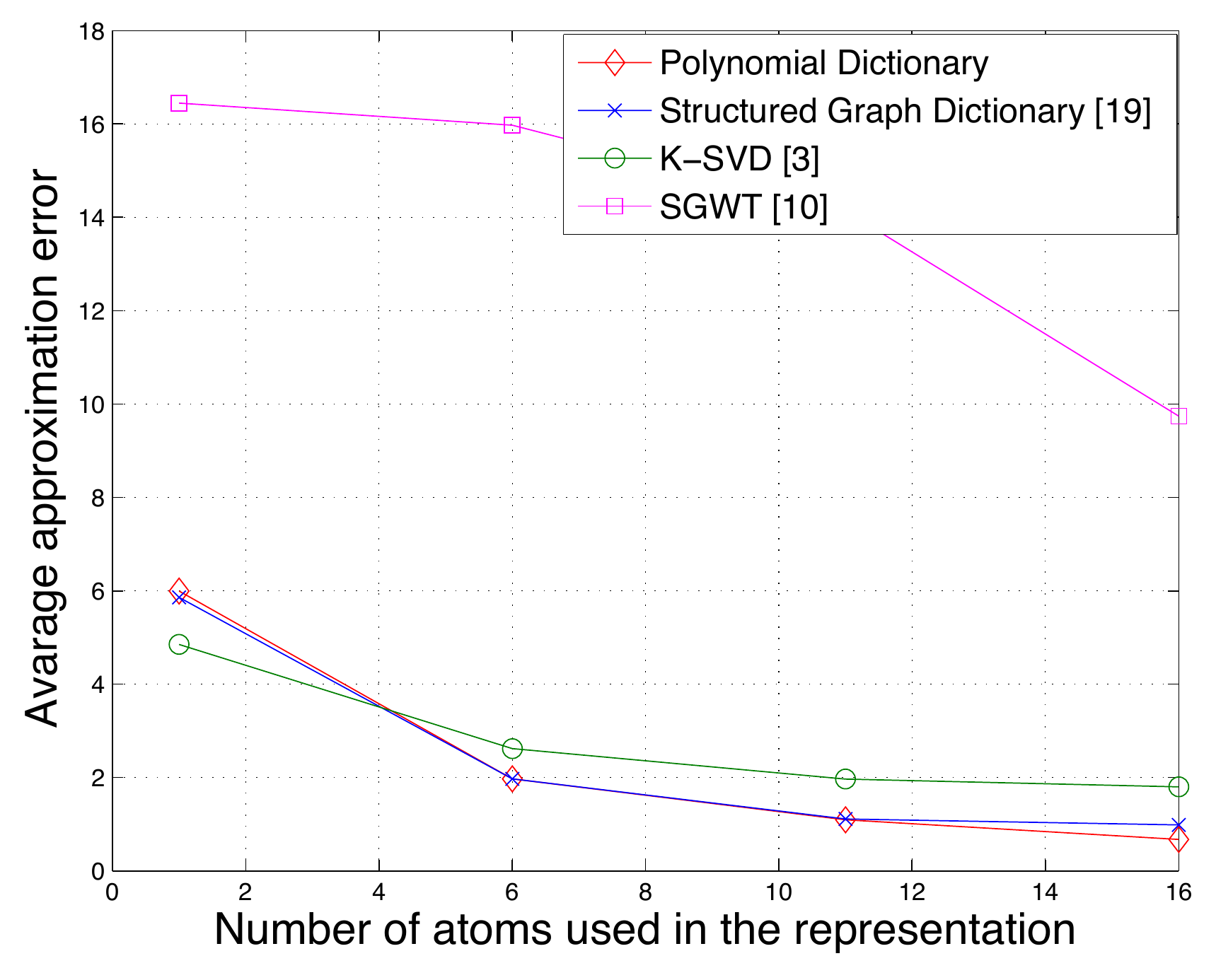}\label{fig:approximation_FickrOMP}}
           \subfigure [Traffic Dataset] { \includegraphics[width=7cm]{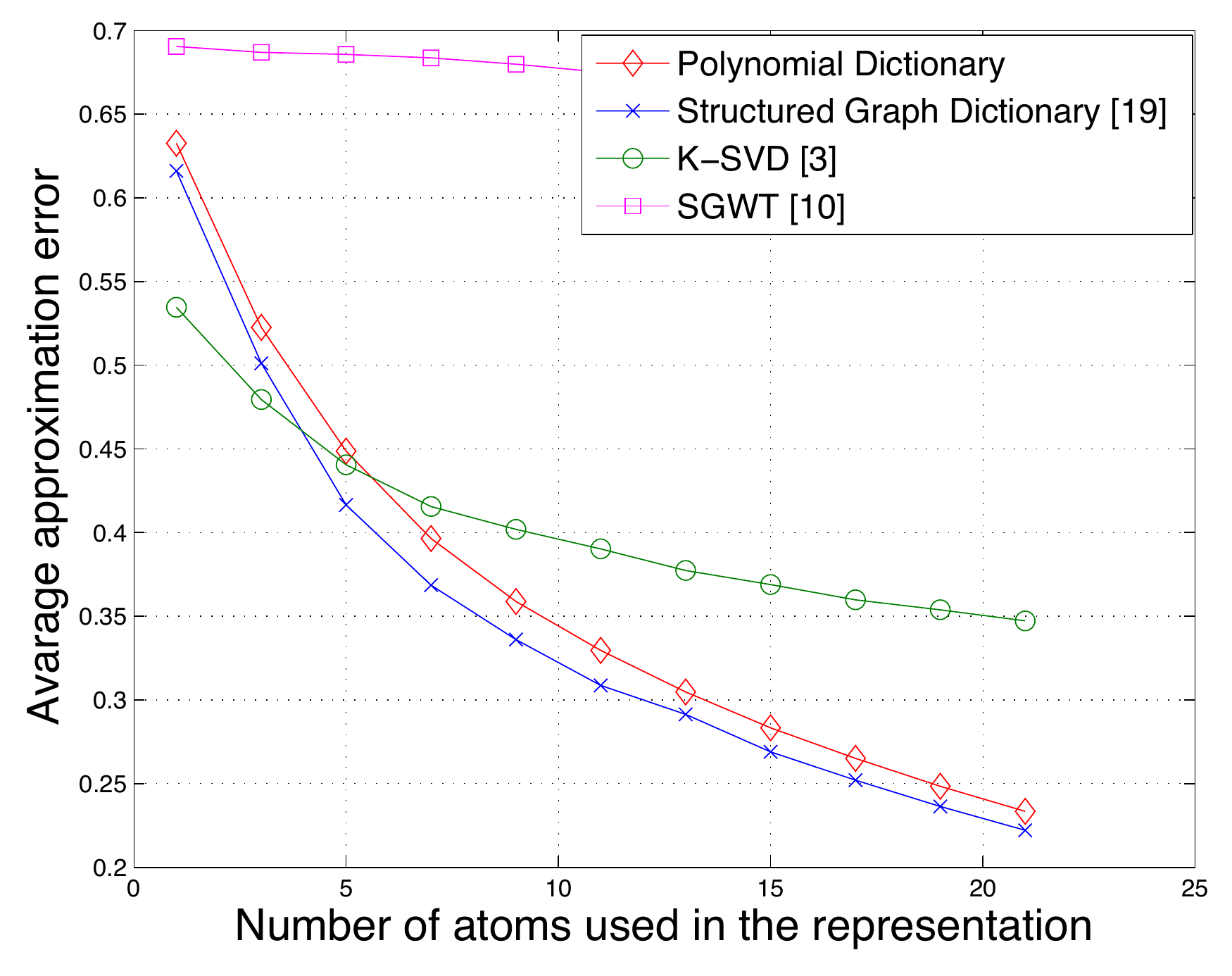}\label{fig:approximation_trafficOMP}}
            \subfigure [Brain Dataset] { \includegraphics[width=7cm]{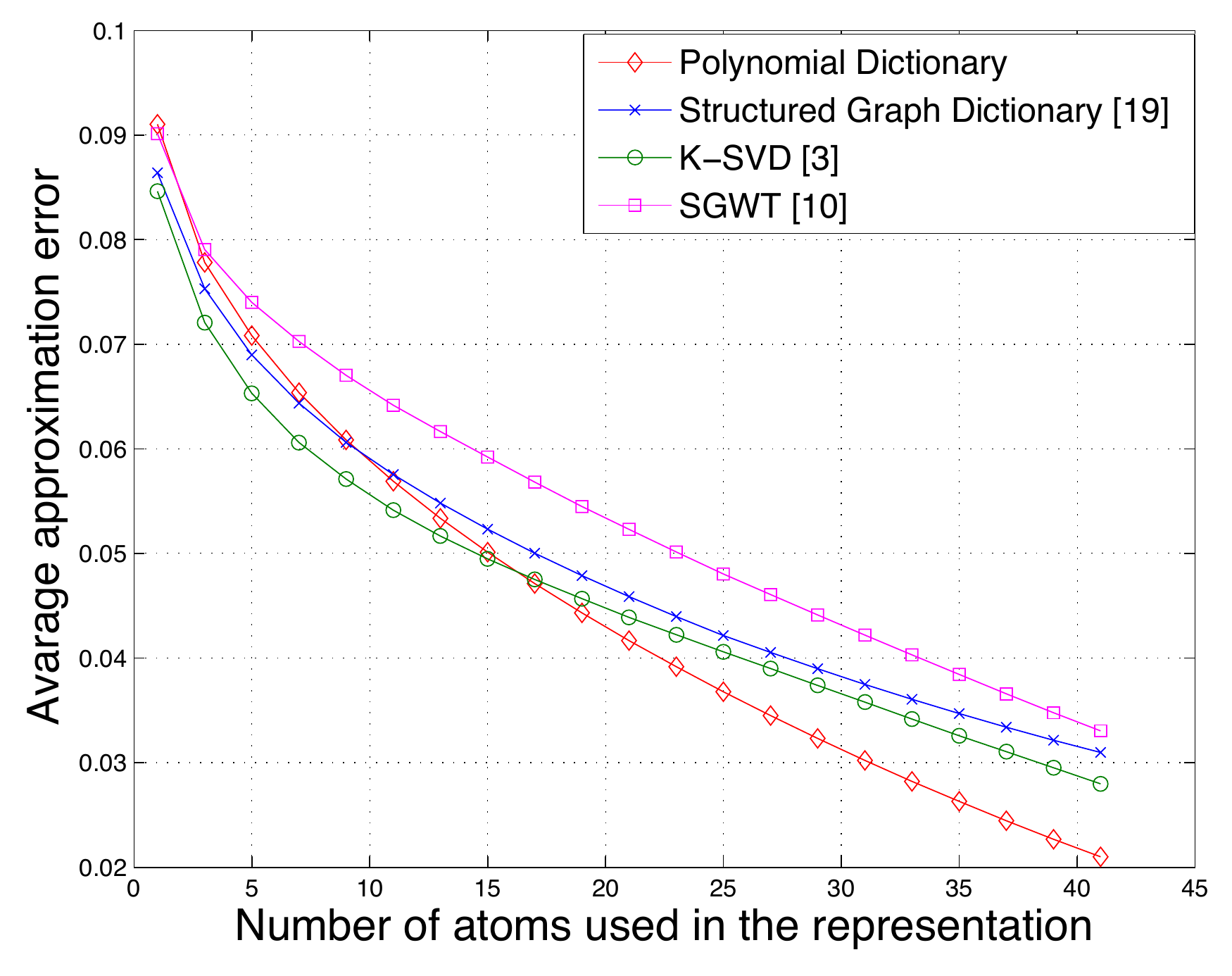}\label{fig:approximation_brainOMP}}
        \caption{
      Comparison of the learned polynomial dictionaries to   the SGWT, K-SVD, and graph structured dictionaries \cite{Zhang2012} in terms of approximation performance on testing data generated from  the (a) Flickr, (b) traffic, and (c) brain datasets, respectively.}
%        Comparison of the polynomial dictionary  with the SGWT, K-SVD and the graph structured dictionary \cite{Zhang2012} in terms of approximation performance in the (a) Flickr, (b) traffic and (c) brain dataset.} 
        \label{fig:flickr_exp}
\vspace{-0.5cm}
\end{figure}

%Examples of these  graph signals  are illustrated in Fig. \ref{fig:dataset_examples}. 
Fig. \ref{fig:flickr_exp} shows the approximation performance of the learned polynomial dictionaries for the three different  datasets.    The  behavior is similar  in all  three datasets, and also similar to results on the synthetic datasets in the previous section. In particular, the data-adapted dictionaries clearly outperform the SGWT dictionary in terms of approximation error on test signals, and the localized atoms of the learned polynomial dictionary effectively represent the real graph signals.

 % More precisely, we observe that the polynomial dictionary again outperforms the SGWT. It can even achieve better performance than K-SVD when sparsity increases. In particular, we observe that K-SVD outperforms both graph structured algorithms for a small sparsity level as it learns atoms that can smoothly approximate the whole signal. 
 
\begin{figure}
      \centering
        % \subfigure [K-SVD] {\includegraphics[width=7.5cm]{Figures/brain_atomsKSVD_most_used}\label{fig:atoms_KSVD}}
        \subfigure [K-SVD] {\includegraphics[width=8cm]{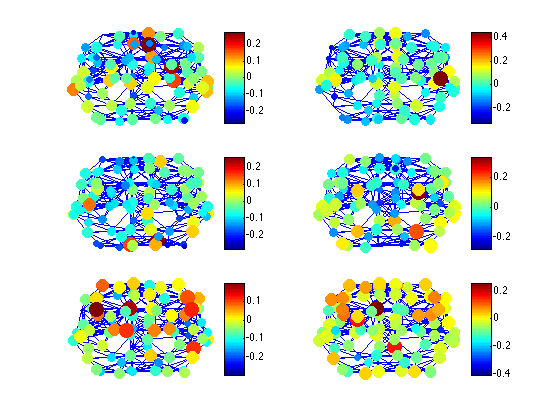}\label{fig:atoms_KSVD}}
        %  \subfigure [K-SVD] {\includegraphics[width=7.5cm]{Figures/atoms_KSVD_3.png}\label{fig:atoms_KSVD}}
           %\subfigure [Polynomial Dictionary] { \includegraphics[width=7.5cm]{Figures/brain_atoms_pol}\label{fig:atoms_pol}}
             \subfigure [Polynomial Dictionary] { \includegraphics[width=8cm]{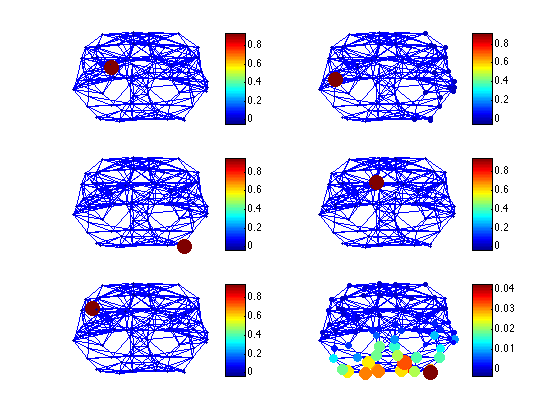}\label{fig:atoms_pol}}
             % \subfigure [K-SVD] {\includegraphics[width=7.5cm]{Figures/atoms_KSVD_most_used.pdf}\label{fig:atoms_KSVD}}
             % \subfigure [K-SVD] {\includegraphics[width=7.5cm]{Figures/brain_atomsKSVD_most_used}\label{fig:atoms_KSVD}}
        \caption{Examples of atoms learned from training data from the brain dataset with (a) K-SVD and (b) the polynomial dictionary. The six atoms are the ones that were most commonly included by OMP in the sparse decomposition of the testing signals. Notice that the atoms in (a) have a support that covers the entire graph while the atoms in (b) are localized around a vertex.}% {\color{red}[How were these 6 atoms selected? The ones that were most commonly included in the sparse set in OMP on the test data?]} } 
        \label{fig:atom_comparison}
\vspace{-0.5cm}
\end{figure}

 In Fig. \ref{fig:atom_comparison}, we illustrate the six most used atoms after applying OMP for the sparse decomposition of the testing signals from the brain dataset in the learned  K-SVD dictionary    and our learned polynomial dictionary.  %In the case of the polynomial dictionary, we choose to plot the atoms from the $S=2$ learned patterns placed on three random vertices on the graph.  Due to the structure of the dictionary, the atoms placed on the other nodes show similar behavior.
Note that in Fig. \ref{fig:atoms_pol}, the polynomial dictionary consists of localized atoms with  supports concentrated on the close neighborhoods of different vertices, which can lead to  poor approximation performance at low sparsity levels. However, as the sparsity level increases, the localization property clearly becomes beneficial. As was the case in the synthetic datasets, the learned polynomial dictionary also has the ability to represent localized patterns that did not appear in the training data, unlike K-SVD.  For example, an unexpected event in  London  could significantly increase the number of pictures taken in the vicinity of a location that was not necessarily popular in the past. 

Comparing our algorithm with the one of  \cite{Zhang2012}, we observe that the performance of the latter is sometimes better (see Fig. \ref{fig:approximation_trafficOMP}) and sometimes worse (see Figs. \ref{fig:approximation_FickrOMP}, \ref{fig:approximation_brainOMP}).  Apart from the differences  between the two algorithms that we have already discussed in the previous subsections, one drawback of  \cite{Zhang2012} is the way the dictionary is updated. Specifically, the update of the dictionary is performed block by block, which  leads to a local optimum in the dictionary update step. This can lead to worse performance when compared to our algorithm, where all subdictionaries are updated simultaneously.
%procedure in the dictionary learning step can deteriorate the performance in comparison to our algorithm, where all the subdictionaries are updated simultaneously. %  The performance of the polynomial dictionary is quite similar to that obtained with \cite{Zhang2012} with the difference that the latter does not have an easily implementable structure. In order to obtained an efficiently  implementable dictionary, we fit a polynomial function to the discrete values $\widehat{g_s}$  learned with \cite{Zhang2012} in the traffic experiments. We observe that our polynomial dictionary outperforms the polynomial approximation of   \cite{Zhang2012}. %,  we observe that this localization property of the dictionary can be beneficial for a sparsity level bigger than four. 

%  \begin{figure*}[t]
%\begin{minipage}{2.1in}
%\begin{center}
%~ \includegraphics[width=1.1\textwidth]{Figures/brain_atomsKSVD}~ \\
%~~~(a) Flickr Dataset ~\\
%\end{center}
%\end{minipage}
%\hfill
%\begin{minipage}{2.1in}
%\begin{center}
%~\includegraphics[width=1.1\textwidth]{Figures/brain_atoms_pol}~\\
%~~~(b) Traffic Dataset ~\\
%\end{center}
%\end{minipage}
%\caption{Examples of  graph signals from three different datasets: (a) Number of distinct Flickr users that have taken photos near the Trafalgar square in one day. (b) Daily bottleneck (in minutes) in the Alameda County. (c) An fMRI signal acquired from a subject.  }  
%\label{fig:dataset_examples}
%\vspace{-0.1cm}
%\end{figure*}
 
 %The next set of experiments consists of some data collected from the Flickr dataset.  Specifically, 
 \section{Computational Efficiency of the Learned Polynomial Dictionary}\label{practical_issues}
 The structural properties of the proposed class of dictionaries lead to compact representations and computationally efficient implementations, which we elaborate on briefly in this section.
First,  the number of free parameters depends on the number $S$ of subdictionaries and the  degree $K$ of the polynomials.  The total number of parameters is $(K+1)S$, and since $K$ and  $S$ are small  in practice, the dictionary is compact and easy to store.  Second, contrary to the non-structured dictionaries learned by algorithms such as K-SVD and MOD, 
the dictionary forward and adjoint operators can be efficiently applied when the graph is sparse, as is usually the case in practice. Recall from (\ref{subdictionary}) that $\D^Ty=\sum_{s=1}^S\sum_{k=0}^K\alpha_{sk}\L^k$. The computational cost of the iterative sparse matrix-vector multiplication required to compute $\{\mathcal{L}^{k}y\}_{ k=1,2,...,K}$ is $O(K|\E|)$. Therefore, the total computational cost  to compute $\D^Ty$ is  $O(K|\E|+NSK)$. We further note that, by following a procedure similar to the one in  \cite[Section 6.1]{Hammond2010}, the term $\D\D^Ty$ can also be computed in a fast way by exploiting the fact that $ \D\D^Ty=\sum_{s=1}^S\widehat{g_s}^2(\mathcal{L})y$. This leads to a polynomial of degree $K^{'}=2K$ that can be efficiently computed.
 Both operators $\D^Ty$ and $\D\D^Ty$ are important components of most sparse coding techniques. 
 
 Therefore, these efficient implementations are  useful in numerous signal processing tasks, and comprise one of the main advantages of learning structured parametric dictionaries. For example, to find sparse representations of different signals with the learned dictionary, rather than using OMP, we can use iterative soft thresholding \cite{daubechies2004} to solve the lasso regularization problem \cite{Tibshirani94}. The two main operations required in iterative soft thresholding, $\D^Ty$ and $\D^T\D x$, can both be approximated by the Chebyshev approximation method of \cite{Hammond2010}, as explained in more detail in \cite[Section IV.C]{Shuman11}.
 
 A third computational benefit is that in settings where the data is distributed and communication between nodes of the graph is costly (e.g., a sensor network), the polynomial structure of the learned dictionary enables quantities such as $\D^Ty$, ${\cal D}x$, $\D\D^Ty$, and $\D^T\D x$ to be efficiently computed in a distributed fashion using the techniques of \cite{Shuman11}. As discussed in \cite{Shuman11}, this structure makes it possible to implement many signal processing algorithms in a distributed fashion.

 \section{Conclusion}
We proposed a parameterized family of structured dictionaries -- namely, unions of polynomial matrix functions of the graph Laplacian -- to sparsely represent signals on a given weighted graph, and an algorithm to learn the parameters of a dictionary belonging to this family from a set of training signals on the graph. 
%that sparsely represent signals on a given weighted graph. 
When translated to a specific vertex, the learned polynomial kernels in the graph spectral domain correspond to localized patterns on the graph. The fact that we translate each of these patterns to different areas of the graph led to sparse approximation performance that was clearly better than that of non-adapted graph wavelet dictionaries such as the SGWT and comparable to or better than that of dictionaries learned by state-of-the art numerical algorithms such as K-SVD. The approximation performance of our learned dictionaries was  also more robust to the available size of training data. At the same time, because our learned dictionaries are unions of polynomial matrix functions of the graph Laplacian, they can be efficiently stored and implemented in both centralized and distributed signal processing tasks.

\section*{{\textsc Acknowledgements}}
The authors would like to thank Prof. Dimitri Van De Ville for providing the brain data, Prof. Antonio Ortega, Prof. Pierre Vandergheynst and Xiaowen Dong for their useful feedbacks about this work,  and Giorgos Stathopoulos for helpful discussions about the optimization problem.  

  \appendices 

\section{}
\label{QP}
The optimization problem (\ref{eq:opt_prob_fix_X}) is  a quadratic program as it consists of a quadratic objective function and a set of affine constraints. In particular, using (\ref{subdictionary}), the objective function can be written as

\begin{equation}
{\small
\begin{split}
|| Y - \D X ||^{2}_{F}+\mu \|\alpha\|_2^2&= \sum_{n=1}^N\sum_{m=1}^M(Y - \D X )_{nm}^2+\mu \alpha^T\alpha \\ &=\sum_{n=1}^N\sum_{m=1}^M\left(Y -\sum_{s=1}^S\sum_{k=0}^K\alpha_{sk}{\L}^{k} X_s \right)_{nm}^2+\mu \alpha^T\alpha,
\end{split}
\label{detailed_expression}
  }
\end{equation}
where $X_s\in \Rbb^{N\times M}$ denotes the rows of the matrix $X$ corresponding to the atoms in the subdictionary $\D_s$. Let us define the column vector $P_{nm}^{s}\in\Rbb^{(K+1)}$ as
\begin{equation*}
\small{
P_{nm} ^ s=[(\L^0)_{(n,:)}{X_s}_{(:,m)}; (\L^1)_{(n,:)}{X_s}_{(:,m)};...;(\L^K)_{(n,:)}{X_s}_{(:,m)}  ], 
}
\end{equation*}
where $(\L^k)_{(n,:)}$ is the $n^{th}$ row of the $k^{th}$ power of the Laplacian matrix and ${X_s}_{(:,m)} $ is the $m^{th}$ column of the matrix $X_s$.
We then stack these column vectors into the column vector $P_{nm}\in\Rbb^{S(K+1)}$, which is defined as $P_{nm}=[P_{nm}^1;P_{nm}^2;...;P_{nm}^{S}]$.
%as %$P^{(nm)}_s\in\Rbb^{(K+1)}$ the vector containing the entries 
%\begin{equation}
%\small{
%P^{(nm)}_s=[\L_{(n,:)}^0{X_s}_{(:,m)}; \L_{(n,:)}{X_s}_{(:,m)};...;\L_{(n,:)}^K{X_s}_{(:,m)}  ] 
%}
%\end{equation}
%
%and 
% $P^{(nm)}\in\Rbb^{S(K+1)}$ the column vector containing the entries 
%\begin{equation*}
%\small{
%P_{nm}=[P_{nm}^1;P_{nm}^2;...;P_{nm}^{S} ].
%}
%%\nonumber
%%\label{Pnm}
%\end{equation*}
%Each $P_{nm}^{s}\in\Rbb^{(K+1)}$ is a column vector whose entries correspond to 
Using this definition of $P_{nm}$, (\ref{detailed_expression}) can be written as 
\begin{equation*}
{\small
\begin{split}
|| Y - \D X ||^{2}_{F}+\mu \|\alpha\|_2^2&= \sum_{n=1}^N\sum_{m=1}^M(Y_{nm} - P_{nm}^T\alpha )^2+\mu \alpha^T\alpha \\ &= \sum_{n=1}^N\sum_{m=1}^M Y_{nm}^2-2Y_{nm} P_{nm}^T\alpha+\alpha^T P_{nm} P_{nm}^T\alpha+\mu \alpha^T\alpha\\
&=\|Y\|_F^2-2\left( \sum_{n=1}^N\sum_{m=1}^MY_{nm}P_{nm}^T\right)\alpha+\alpha^T\left(\sum_{n=1}^N\sum_{m=1}^MP_{nm}P_{nm}^T+\mu I_{S(K+1)}\right)\alpha,
\end{split}
%\label{detailed_expression2}
  }
\end{equation*}
where $I_{S(K+1)}$ is the $S(K+1)\times S(K+1)$ identity matrix. The matrix $\sum_{n=1}^N\sum_{m=1}^MP_{nm}P_{nm}^T+\mu I$ is positive semidefinite, which implies that our objective is quadratic. 

Finally, the optimization  constraints can be %easily 
expressed as affine functions with
$$0 \leq I_S\otimes B\alpha \leq c\boldsymbol{1}$$ and
$$(c-\epsilon)\boldsymbol{1} \leq \boldsymbol{1}^T\otimes B\alpha \leq (c+\epsilon)\boldsymbol{1}~,$$
where the inequalities are component-wise inequalities, $\boldsymbol{1}$ is the vector of ones, $I_S$ is the $S\times S$ identity matrix, and $B$ is the Vandermonde matrix  
$$B=\begin{bmatrix}	
        1 & \lambda_0&\lambda_0^2\dots \lambda_0^K \\
        1 & \lambda_1&\lambda_1^2\dots \lambda_1^K \\ \vdots &\vdots&\vdots\\    1 & \lambda_{N-1}&\lambda_{N-1}^2\dots \lambda_{N-1}^K \\\end{bmatrix}.$$
        Thus, the optimization problem is quadratic and it can be solved  efficiently.

%{\color{red} Other things to take care of:
%\begin{itemize}
%%\item Footnote on front page saying that part of this work will be presented at GlobalSIP
%%\item Funding acknowledgements
%%\item Larger fonts on figures, no horizontal bars on bottom, and smaller figure sizes (in kilobytes)
%\item Double column version (must be 13 pages or less)
%\end{itemize}
%} 

\bibliographystyle{IEEEbib}
\bibliography{mybibfile}

\begin{thebibliography}{10}

\bibitem{Shuman13}
D.~I Shuman, S.~K. Narang, P.~Frossard, A.~Ortega, and P.~Vandergheynst,
\newblock ``The emerging field of signal processing on graphs: Extending
  high-dimensional data analysis to networks and other irregular domains,''
\newblock {\em IEEE Signal Process. Mag.}, vol. 30, no. 3, pp. 83--98, May
  2013.

\bibitem{Rubinstein2010overview}
R.~Rubinstein, A.~M. Bruckstein, and M.~Elad,
\newblock ``Dictionaries for sparse representation modeling,''
\newblock {\em Proc. of the IEEE}, vol. 98, no. 6, pp. 1045 --1057, Apr. 2010.

\bibitem{Aharon06}
M.~Aharon, M.~Elad, and A.~Bruckstein,
\newblock ``{K-SVD}: An algorithm for designing overcomplete dictionaries for
  sparse representation,''
\newblock {\em IEEE Trans. Signal Process.}, vol. 54, no. 11, pp. 4311--4322,
  2006.

\bibitem{Engan99}
K.~Engan, S.~O. Aase, and J.~H. Husoy,
\newblock ``{ Method of optimal directions for frame design},''
\newblock in {\em Proc. IEEE Int. Conf. Acc., Speech, and Signal Process.},
  Washington, DC, USA, 1999, vol.~5, pp. 2443--2446.

\bibitem{Mallat2008}
S.~Mallat,
\newblock {\em A Wavelet Tour of Signal Processing: The Sparse Way},
\newblock Academic Press, 3rd edition, 2008.

\bibitem{Jost2006}
P.~Jost, P.~Vandergheynst, S.~Lesage, and R.~Gribonval,
\newblock ``Motif: An efficient algorithm for learning translation invariant
  dictionaries,''
\newblock in {\em Proc. IEEE Int. Conf. Acc., Speech, and Signal Process.},
  Toulouse, France, 2006, vol.~5, pp. 857--860.

\bibitem{Yaghoobi2009}
M.~Yaghoobi, L.~Daudet, and M.~E. Davies,
\newblock ``Parametric dictionary design for sparse coding,''
\newblock {\em IEEE Trans. Signal Process.}, vol. 57, no. 12, pp. 4800--4810,
  Dec. 2009.

\bibitem{Rubinstein2010}
R.~Rubinstein, M.~Zibulevsky, and M.~Elad,
\newblock ``Double sparsity: learning sparse dictionaries for sparse signal
  approximation,''
\newblock {\em IEEE Trans. Signal Process.}, vol. 58, no. 3, pp. 1553--1564,
  2010.

\bibitem{ThanouParamDL}
D.~Thanou, D.~I Shuman, and P.~Frossard,
\newblock ``{Parametric dictionary learning for graph signals},''
\newblock in {\em Proc. IEEE Glob. Conf. Signal and Inform. Process.}, Austin,
  Texas, Dec. 2013.

\bibitem{Hammond2010}
D.~Hammond, P.~Vandergheynst, and R.~Gribonval,
\newblock ``Wavelets on graphs via spectral graph theory,''
\newblock {\em Appl. Comput. Harmon. Anal.}, vol. 30, no. 2, pp. 129--150,
  March 2010.

\bibitem{Zhu12}
X.~Zhu and M.~Rabbat,
\newblock ``Approximating signals supported on graphs,''
\newblock in {\em Proc. IEEE Int. Conf. Acc., Speech, and Signal Process.},
  Kyoto, Japan, Mar. 2012, pp. 3921--3924.

\bibitem{Coifman06}
R.~R. Coifman and M.~Maggioni,
\newblock ``Diffusion wavelets,''
\newblock {\em Appl. Comput. Harmon. Anal.}, vol. 21, pp. 53--94, March 2006.

\bibitem{NarangO12}
S.~K. Narang and A.~Ortega,
\newblock ``Perfect reconstruction two-channel wavelet filter banks for graph
  structured data,''
\newblock {\em IEEE Trans. Signal Process.}, vol. 60, no. 6, pp. 2786--2799,
  June 2012.

\bibitem{ShumanWindGFT}
D.~I Shuman, B.~Ricaud, and P.~Vandergheynst,
\newblock ``{A windowed graph Fourier transform},''
\newblock in {\em Proc. IEEE Stat. Signal Process. Wkshp.}, Michigan, Aug.
  2012.

\bibitem{shuman_ACHA_2013}
D.~I Shuman, B.~Ricaud, and P.~Vandergheynst,
\newblock ``Vertex-frequency analysis on graphs,''
\newblock {\em submitted to Appl. Comput. Harmon. Anal.}, July 2013.

\bibitem{shuman_TSP_2013}
{D. I Shuman}, Christoph Wiesmeyr, Nicki Holighaus, and Pierre Vandergheynst,
\newblock ``Spectrum-adapted tight graph wavelet and vertex-frequency frames,''
\newblock {\em submitted to IEEE Trans. Signal Process.}, Nov. 2013.

\bibitem{bremer_packets}
J.~C. Bremer, R.~R. Coifman, M.~Maggioni, and A.~D. Szlam,
\newblock ``Diffusion wavelet packets,''
\newblock {\em Appl. Comput. Harmon. Anal.}, vol. 21, no. 1, pp. 95--112, 2006.

\bibitem{Rustamov2013}
R.~M. Rustamov and L.~Guibas,
\newblock ``Wavelets on graphs via deep learning,''
\newblock in {\em Advances in Neural Information Processing Systems (NIPS)},
  2013.

\bibitem{Zhang2012}
X.~Zhang, X.~Dong, and P.~Frossard,
\newblock ``{Learning of structured graph dictionaries},''
\newblock in {\em Proc. IEEE Int. Conf. Acc., Speech, and Signal Process.},
  Kyoto, Japan, Mar. 2012, pp. 3373 -- 3376.

\bibitem{higham}
N.~J. Higham,
\newblock {\em Functions of Matrices},
\newblock Society for Industrial and Applied Mathematics, 2008.

\bibitem{MiaoZheng2011}
M.~Zheng, J.~Bu, C.~Chen, C.~Wang, L.~Zhang, G.~Qiu, and D.~Cai,
\newblock ``Graph regularized sparse coding for image representation,''
\newblock {\em IEEE Trans. Image Process.}, vol. 20, no. 5, pp. 1327--1336, May
  2011.

\bibitem{Chung97}
F.~Chung,
\newblock {\em Spectral Graph Theory},
\newblock American Mathematical Society, 1997.

\bibitem{Tropp04}
J.~A. Tropp,
\newblock ``Greed is good: Algorithmic results for sparse approximation,''
\newblock {\em IEEE Trans. Inform. Theory}, vol. 50, no. 10, pp. 2231--2242,
  2004.

\bibitem{Bruckstein2009}
M.~Bruckstein, D.~Donoho, and M.~Elad,
\newblock ``From sparse solutions of systems of equations to sparse modeling of
  signals and images,''
\newblock {\em SIAM Rev.}, vol. 51, no. 1, pp. 34--81, Feb. 2009.

\bibitem{Elad2010}
M.~Elad,
\newblock {\em Sparse and Redundant Representations - From Theory to
  Applications in Signal and Image Processing},
\newblock Springer, 2010.

\bibitem{BoydConvex}
S.~Boyd and L.~Vandenberghe,
\newblock {\em Convex Optimization},
\newblock New York: Cambridge University Press, 2004.

\bibitem{Boyd_admm}
S.~Boyd, N.~Parikh, E.~Chu, B.~Peleato, and J.~Eckstein,
\newblock ``Distributed optimization and statistical learning via the
  alternating direction method of multipliers,''
\newblock {\em Foundations and Trends in Machine Learning}, vol. 3, no. 1, pp.
  1--122, 2011.

\bibitem{SDPT3}
K.C. Toh, M.J. Todd, and R.H. Tutuncu,
\newblock ``{SDPT3} --- a {MATLAB} software package for semidefinite
  programming,''
\newblock in {\em Optimization Methods and Software}, 1999, vol.~11, pp.
  545--581.

\bibitem{YALMIP}
J.~Lofberg,
\newblock ``{YALMIP}: A toolbox for modeling and optimization in {MATLAB},''
\newblock in {\em Proc. CACSD Conf.}, Taipei, Taiwan, 2004.

\bibitem{Dong2013}
X.~Dong, A.~Ortega, P.~Frossard, and P.~Vandergheynst,
\newblock ``{Inference of mobility patterns via spectral graph wavelets},''
\newblock in {\em Proc. IEEE Int. Conf. Acc., Speech, and Signal Process.},
  Vancouver, Canada, May 2013.

\bibitem{Choe2002}
T.~Choe, A.~Skabardonis, and P.~P. Varaiya,
\newblock ``{Freeway performance measurement system (PeMS): an operational
  analysis tool},''
\newblock in {\em Presented at Annual Meeting of Transportation Research
  Board}, Washington, DC, USA, Jan. 2002.

\bibitem{EryilmazVSV11}
H.~Eryilmaz, D.~Van~De Ville, S.~Schwartz, and P.~Vuilleumier,
\newblock ``Impact of transient emotions on functional connectivity during
  subsequent resting state: A wavelet correlation approach,''
\newblock {\em NeuroImage}, vol. 54, no. 3, pp. 2481--2491, 2011.

\bibitem{RichiardiESVV11}
J.~Richiardi, H.~Eryilmaz, S.~Schwartz, P.~Vuilleumier, and D.~Van~De Ville,
\newblock ``Decoding brain states from {fMRI} connectivity graphs,''
\newblock {\em NeuroImage}, vol. 56, no. 2, pp. 616--626, 2011.

\bibitem{daubechies2004}
I.~Daubechies, M.~Defrise, and C.~De Mol,
\newblock ``An iterative thresholding algorithm for linear inverse problems
  with a sparsity constraint,''
\newblock {\em Commun. Pure Appl. Math.}, vol. 57, no. 11, pp. 1413--1457, Nov.
  2004.

\bibitem{Tibshirani94}
R.~Tibshirani,
\newblock ``Regression shrinkage and selection via the lasso,''
\newblock {\em J. R. Stat. Soc. Ser. B}, vol. 58, pp. 267--288, 1994.

\bibitem{Shuman11}
D.~I Shuman, P.~Vandergheynst, and P.~Frossard,
\newblock ``Chebyshev polynomial approximation for distributed signal
  processing,''
\newblock in {\em Proc. Int. Conf. Distr. Comput. Sensor Sys.}, Barcelona,
  Spain, June 2011.

\end{thebibliography}

\end{document}